\documentclass[letterpaper, 10pt, conference]{ieeeconf}      
\IEEEoverridecommandlockouts                              

\usepackage{graphicx} 
\usepackage{times} 
\usepackage{amsmath} 
\usepackage{bm}
\usepackage{breqn}
\usepackage{ragged2e}
\usepackage{xfrac}
\usepackage{amssymb} 
\usepackage{algorithm}
\usepackage[noend]{algpseudocode}
\usepackage{multirow,color}
\usepackage{cite}
\usepackage{algpseudocode}
\usepackage{varwidth}
\usepackage{subfig}
\usepackage{booktabs}
\usepackage{gensymb}
\usepackage[hyphens]{url}
\usepackage[export]{adjustbox}
\usepackage[font=footnotesize]{caption}
\let\labelindent\relax
\usepackage[inline]{enumitem} 
\usepackage{dblfloatfix}
\usepackage{tabularx}
\usepackage{setspace}
\usepackage{bm}
\usepackage{textcomp}
\usepackage{multirow}
\usepackage{makecell}
\usepackage{comment}

\usepackage{color}
\usepackage{graphicx,tipa}

\makeatletter
\def\BState{\State\hskip-\ALG@thistlm}
\makeatother

\title{\LARGE \bf A Comprehensive General Model of Tendon-Actuated Concentric Tube Robots with Multiple Tubes and Tendons}

\author{Pejman Kheradmand$^{1}$, Behnam Moradkhani$^{1}$, Raghavasimhan Sankaranarayanan$
^{1}$, Kent K. Yamamoto$^{2}$,\\ Tanner J. Zachem$^{2}$, Patrick J. Codd$^{2}$, Yash Chitalia$^{1}$ and Pierre E. Dupont$^{3}$, \textit{Fellow, IEEE} 
\thanks{$^{1}$P. Kheradmand, B. Moradkhani, R. Sankaranarayanan and Y. Chitalia are with the Healthcare Robotics and Telesurgery Laboratory (Heartlab), University of Louisville, Louisville, KY, USA.}%
\thanks{$^{2}$K. K. Yamamoto, T. J. Zachem, P. J. Codd are with Mechanical Engineering and Materials Science, Duke University, Durham, NC, USA.}%
\thanks{$^{3}$ P. E. Dupont is with the Department of Cardiac Surgery, Boston Children’s Hospital, Harvard Medical School, Boston, MA, USA.}%
\thanks{Corresponding Author: P. Kheradmand ({\tt \small pejman.kheradmand@louisville.edu})}
\thanks{This work was supported in part by the NIH under grant R01HL124020 and in part by the Kentucky Spinal Cord and Head Injury Research Trust Fund.}
}

\begin{document}
\maketitle
\thispagestyle{empty}
\begin{abstract}
Tendon-actuated concentric tube mechanisms combine the advantages of tendon-driven continuum robots and concentric tube robots while addressing their respective limitations. They overcome the restricted degrees of freedom often seen in tendon-driven designs, and mitigate issues such as snapping instability associated with concentric tube robots. However, a complete and general mechanical model for these systems remains an open problem. In this work, we propose a Cosserat rod-based framework for modeling the general case of \( n \) concentric tubes, each actuated by \( m_i \) tendons, where \( i = \{1, \ldots, n\} \). The model allows each tube to twist and elongate while enforcing a shared centerline for bending.
We validate the proposed framework through experiments with two-tube and three-tube assemblies under various tendon routing configurations, achieving tip prediction errors $<4\%$ of the robot's total length. We further demonstrate the model’s generality by applying it to existing robots in the field, where maximum tip deviations remain around $5\%$ of the total length. This model provides a foundation for accurate shape estimation and control of advanced tendon-actuated concentric tube robots.

\end{abstract}
\section{Introduction}\label{sec:introduction}
Minimally invasive surgical interventions have revolutionized modern medicine by reducing patient trauma, shortening recovery times, and improving procedural outcomes. However, accessing deep-seated anatomical targets, such as the spine, brain, or vasculature, poses significant challenges due to the confined, and deformable nature of biological tissues. While highly accurate in structured environments, traditional rigid-link robotic systems often lack the flexibility and compliance required to safely navigate these constrained anatomical spaces. To overcome these limitations, the field of surgical robotics has increasingly turned toward \textit{continuum robots}, a class of devices characterized by their highly deformable, jointless structures capable of smooth bending and continuous curvature~\cite{walker2016snake}. They have common applications in cardiac interventions \cite{ding2025continuum, chitalia2023model, jeong2020design, nayar2021design, nayar2022toward, qi2023telerobotic}, neurosurgery \cite{price2023using}, abdominal interventions, among other applications~\cite{gosline2012percutaneous, Jessica_Steerable_Cannula, SJTU_Single_Port, Jessica_A_Survey}.

Among the most mature continuum robot architectures are \textit{concentric tube robots (CTRs)}~\cite{dupont2009design} consisting of multiple super-elastic pre-curved tubes nested within each other. The tubes are independently rotated and translated at the base, and their geometric and elastic interactions produce complex spatial configurations~\cite{gilbert2016ctr, rucker2010geometrically}.
Models for pre-curved concentric tube robots \cite{dupont2009design, rucker2010geometrically} have been based on Cosserat rod theory~\cite{antman1995special}, however, they neglect extensibility and transverse shear, simplifying the model to a Kirchhoff’s elastic rod~\cite{dill1992kirchhoff}.
Given that these CTR models do not incorporate tendon actuation, these assumptions are reasonable, as extensibility and transverse shear have minimal influence on the deformation behavior of purely pre-curved concentric tubes. While these foundational models accurately capture bending and torsion for classical CTRs, they have not addressed a more general system with tendon routing and extensibility. Though CTRs provide a compact and dexterous platform for many surgical procedures, their capabilities are inherently constrained by their reliance on tube pre-curvature and base actuation alone. This often results in limitations such as snapping instabilities under external loading conditions~\cite{ha2015elastic, gilbert2015elastic, hendrick2015designing, till2020dynamic}. Snapping instability occurs when a concentric tube robot configuration is not elastically stable, often due to accumulated elastic potential energy from bending and twisting being suddenly released, especially under high pre-curvature, tube stiffness, or external loading~\cite{ha2015elastic}. Over the years, numerous studies on the elastic instabilities of CTRs have been conducted. Works such as~\cite{ha2014achieving, azimian2014structurally, luo2021design} attempted to minimize elastic instability through structural design changes while~\cite{Bergeles_Planning, Leibrandt_2017, bergeles2015concentric} used path planning and constrained optimizations to avoid it. Leibrandt et. al.~\cite{leibrandt_2015} used multi-core computer architectures to compute inverse kinematics in real-time for CTRs while avoiding instabilities and anatomical collisions. While most works try to minimize or avoid these instabilities, Riojas et. al~\cite{riojas2018can} used the elastic instability of CTRs to benefit surgical tasks such as sutures --- leveraging the large amounts of energy that CTRs release during the instability. 

More recent efforts have sought to extend CTR models to include tendon actuation. Chitalia et al.~\cite{chitalia2023modeling} used a two-tube tendon-actuated catheter for transseptal navigation. Their approach includes axial and torsional deformation but exhibits extensibility simplifications that limits the scope to linear tendon routing and cannot be extended to multi-tube, multi-tendon systems.
\begin{figure}
\centering 
\includegraphics[width=\columnwidth]{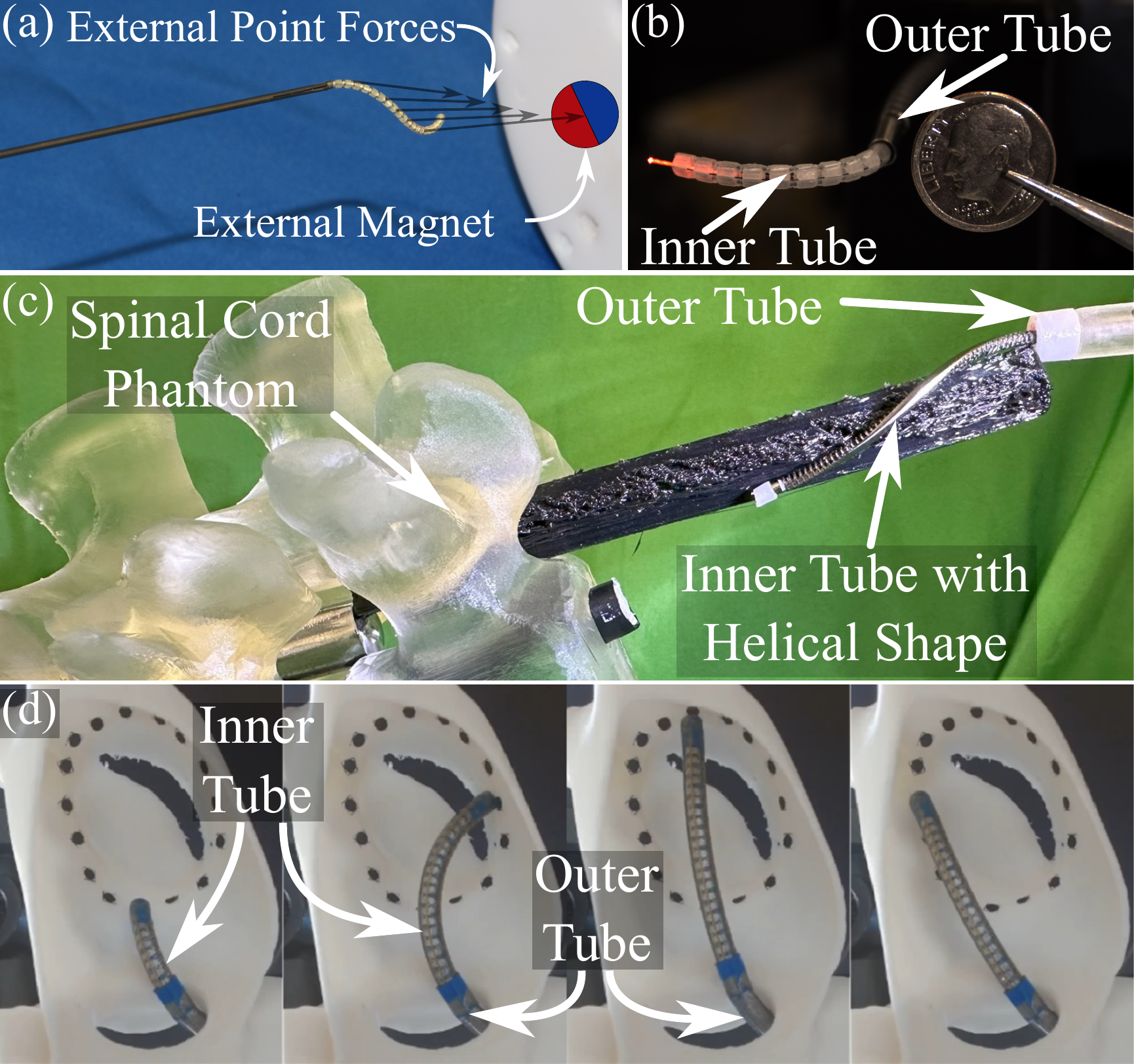} 
\caption{ Various examples of tendon-actuated concentric tube (TACT) robot architectures: (a) Tendon-Assisted Magnetically Steered (TAMS) robotic stylet for brachytherapy\cite{kheradmand2024tams}, (b) Tendon-actuated Concentric Tube Endonasal Robot (TACTER)~\cite{yamamoto2025tacter}, (c) Tendon-actuated concentric tube robot for lateral and ventral spinal cord stimulation\cite{moradkhani2025exonav}, and (d) Telescoping two-tube tendon-actuated catheter for transseptal navigation\cite{chitalia2023modeling}.}
\label{fig:ConcentricRobots} 
\vspace{-5 mm}
\end{figure}
Figure~\ref{fig:ConcentricRobots} shows several tendon actuated concentric tube robot (TACTR) designs developed for different surgical applications. 
These include TAMS~\cite{kheradmand2024tams} for brachytherapy, TACTER~\cite{yamamoto2025tacter} for endonasal procedures, and EXONAV~\cite{moradkhani2025exonav} for spinal cord stimulation. While effective in their implementations, the associated models were highly task-specific, restricted to at most two tubes, and did not capture variable routing effects.
Dupont et al.~\cite{dupont2009design} modeled concentric-tube robots by incorporating bending interactions across multiple pre-curved tubes. However, they assume the shear and axial strains are negligible in their model. Rucker and Webster~\cite{rucker2011statics} et al. developed a general tendon-routing model for a single-tube continuum robot that accounts for shear and elongation. However, their work was limited to single-tube designs.
Grassmann et al.~\cite{grassmann2022dataset} introduced a large-scale dataset and benchmark for learning-based modeling of a three-tube concentric robot. While their shallow neural network achieved higher accuracy than traditional physics-based models in predicting forward kinematics, their approach is purely data-driven and does not offer physical interpretability or extendability to dynamic or inverse problems.
Hachen et al.~\cite{hachen2025nonlinear} developed a model predictive controller (MPC) for tendon-driven continuum robots using a piecewise constant curvature (PCC) model, enabling shape constraint satisfaction and real-time obstacle avoidance. However, the approach is designed for tendon-driven single-backbone robots and is not directly applicable to concentric tube architectures or scenarios involving significant external forces.

Modeling and control of TACTRs remain challenging despite their promise. Most existing models in the literature either restrict themselves to simple configurations (e.g., two tubes with a single tendon per tube), or rely on simplifying assumptions such as \textit{in-extensibility}, \textit{zero shear}, or \textit{static tendon routing}. 
In this work, we address these gaps by proposing a \textit{generalized mechanical model} for TACTRs based on the full Cosserat rod formulation. Our model supports an arbitrary number of nested concentric tubes (indicated by the variable $n$), actuated by an arbitrary number of tendons per tube ($m_i$ tendons, $\forall i \in [1,\dots,n]$ tubes), and allows for both static and parametric tendon routing profiles. Unlike prior efforts, we do not rely on in-extensibility or shear-neglecting assumptions. Instead, we derive a complete set of coupled ordinary differential equations (ODEs) that govern the robot’s shape under internal and external loading conditions, while enforcing compatibility constraints across tubes. This includes a unified treatment of relative twist, axial dilation, tendon routing effects, and distributed tendon-induced moments. 

The contributions of this paper are as follows: 
\begin{itemize}
    \item A comprehensive modeling framework for tendon actuated concentric tube robots comprising an arbitrary number of tubes (indicated by $n$), each actuated by an arbitrary number of tendons per tube (indicated by $m_i$, where $i$ is the tube number). This general formulation enables broad applicability to complex multi-tube, multi-tendon systems regardless of their initial pre-curvature.
    \item No simplifying assumptions (unlike in related works) such as in-extensibility or negligible shear, making it more suitable for accurately capturing the deformations in tendon actuated concentric tube robots.
\end{itemize}

The remainder of this paper is organized as follows: Section~\ref{sec:robot_model} presents the proposed mechanical model. Section~\ref{sec:Experimental_Model_Validation} describes the experimental setup and provides validation of the model through a series of experiments involving two-tube and three-tube configurations, including cases with non-static tendon routing. Tendon assignment strategies and hysteresis effects are also discussed. We further demonstrate the model's generality by validating it on a conventional concentric tube robot (CTR) and multiple tendon-actuated designs reported in the literature. Section~\ref{sec:Limit} outlines limitations and directions for future work. Finally, Section~\ref{sec:conclusion} concludes the paper.

\section{Mechanics-based Model}\label{sec:robot_model}
Our derivation in this section builds upon the modeling frameworks proposed by Rucker et al.~\cite{rucker2010geometrically, rucker2011statics} and Chitalia et al.~\cite{chitalia2023modeling}. We also reconcile the notational and modeling conventions used by Gazzola et al. in~\cite{gazzola2018soft} with the formulation presented in~\cite{rucker2010geometrically}, ensuring consistency throughout our approach.

\subsection{Tube Kinematics, Relative Twist and Variable Dilation}
The differential equation describing the centerline, $\bm{p_{c}(s)} \in \mathbb{R}^3$ of any tube or rod as a function of arc length $s$ is given by:
\begin{align}\label{eq:shape_evol}
    \bm{\dot{p}_{c}(s)} = R(s)\bm{v(s)} \quad \dot{R}(s) = R(s)[\bm{u(s)}]
\end{align}
The vectors $\bm{v}(s)$ and $\bm{u}(s)$ describe how the position and material frame $R(s) \in \mathrm{SO}(3)$ of the rod evolve along the arc length $s$, where $s \in [0, l]$ and $l$ is the total length of the rod.
The $[\,\cdot\,]$  operator indicates conversion of an element of $\mathbb{R}^3$ to its corresponding element in $\mathfrak{so}(3)$ (the Lie algebra of $\mathrm{SO}(3)$). 
Derivatives with respect to arc length are denoted by $\dot{q} = \sfrac{dq}{ds}$. In the following, dependence on arc length is dropped for brevity, e.g., $q(s)$ is written as $q$. All vector quantities are denoted in boldface throughout this work.

\begin{figure}[t]
\centering 
\includegraphics[width=\columnwidth]{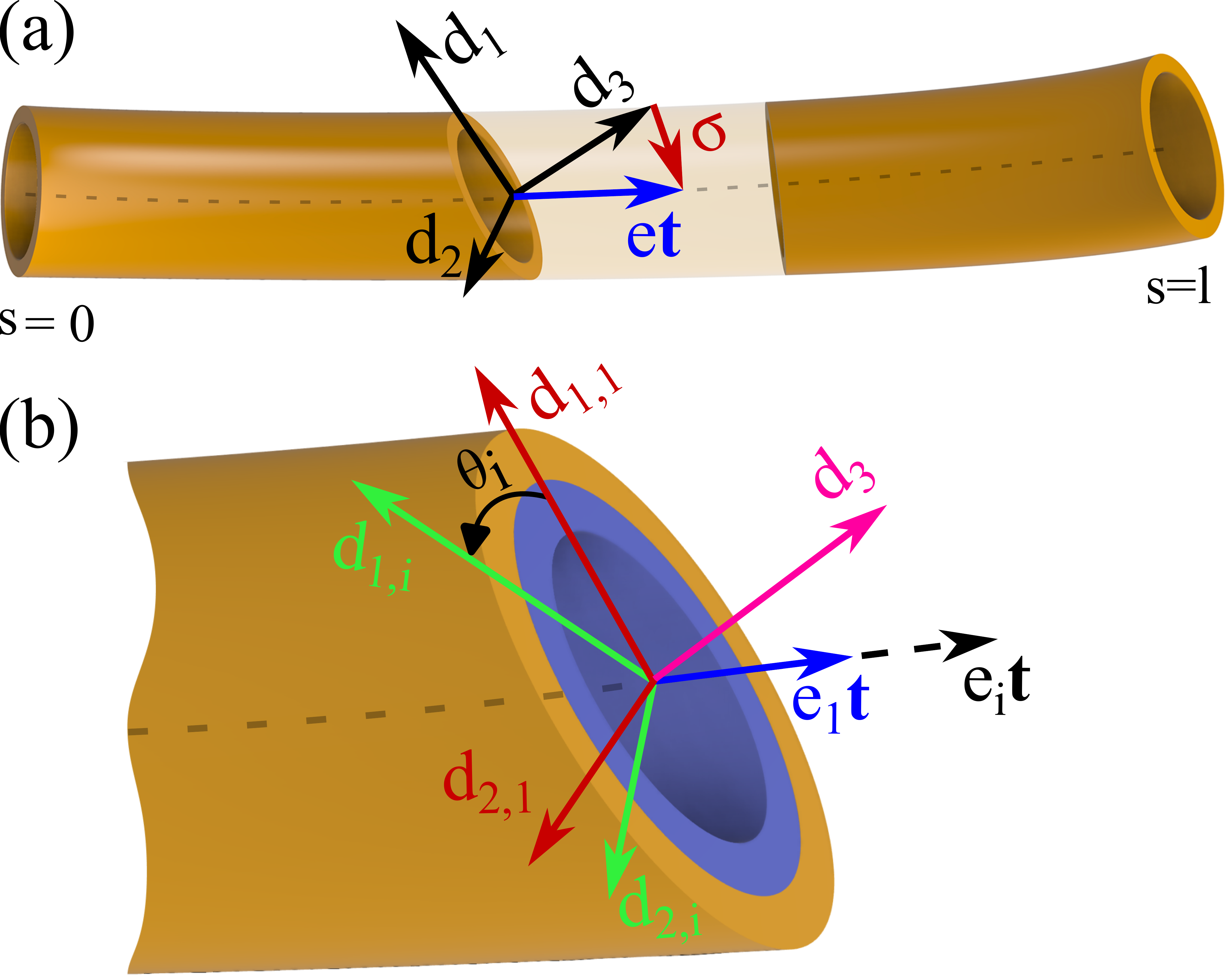} 
\caption{Variable twist and elongation in concentric tubes. (a) A single tube undergoing axial elongation and twist, resulting in a local material frame that dilates and rotates along the arc length. (b) Cross-section of nested concentric tubes illustrating the relative twist ($\theta_i$) and dilation ($e_i$) of an inner tube ($i_{th}$ tube) with respect to the outermost tube, shown through the misalignment of local material frames.}
\label{fig:dilation_twist}
\vspace{-4 mm}
\end{figure}

To interpret the strain vectors geometrically, we examine the relationship between the material frame and the centerline. In the presence of shear and axial strain, cross-sections may no longer remain perpendicular to the centerline (see Fig.~\ref{fig:dilation_twist}(a)), an effect that must be captured to accurately model the robots. The tangent to the centerline is given by $\bm{t}$, while the 
material frame of the tube is defined by the orthonormal directors $[\bm{d_{1}}, \bm{d_{2}}, \bm{d_{3}}]$, which form the columns of the material orientation $R$. Note that $\bm{d_{3}}$ is always perpendicular to the cross-section. Two conventions are considered. Gazzola et al. \cite{gazzola2018soft} define $\bm{\sigma}$ as the shear and axial strains arising from the deviation between the scaled tangent vector $e\bm{t}$ and the local material frame axis \( \bm{d}_3 \):  
\begin{align}
    \bm{\sigma} = R^T(e\bm{t} - \bm{d}_3)
\end{align}
\noindent Here, $e=\frac{ds}{ds^*}$, where $s^*$ is the arc length at rest and $s$ is the arc length when deformed, represents `dilation factor' of the Cosserat rod. Using this dilation factor for a single rod, we express the linear strain vector $\bm{v}$ (from Eq.~(\ref{eq:shape_evol})) as defined in Rucker et al. \cite{rucker2011statics} as:
\begin{align}\label{eq:linear_v}
    \bm{v} = R^T(e\bm{t})
\end{align}
These expressions are consistent, with $\bm{v} = \bm{\sigma} + \mathbb{I}_3$. where $\mathbb{I}_3 = [0,0,1]^T$. Note that $\bm{\sigma}$, $\bm{v}$, and $\bm{u}$ are always expressed in the tube material frame. Rucker et al.\cite{rucker2010geometrically} assumed in-extensibility and zero transverse shear strain in their concentric tube model, leading to $\bm{\sigma} = \mathbf{0}$ and $\bm{v} = \mathbb{I}_3$. This simplification is reasonable for pre-curved tubes without internal actuation. However, in tendon-driven systems, tendon tensions introduce shear and axial loads along the structure. In our work, we retain the full expression for $\bm{\sigma}$ to account for these effects and accurately model TACTRs. In the case of a general $n$-tube TACTR, the third director, which is perpendicular to all the material cross-sections, is consistent across all tubes, i.e., $\bm{d_{3}} = \bm{d_{3,i}}, \ \forall i={1,\dots, n}$. Therefore, the material orientation of tube-$i$ with respect to a global frame is given by $R_i = [\bm{d_{1,i}}, \bm{d_{2,i}}, \bm{d_{3}}]$. However, since the tubes share $\bm{d_{3}}$, the individual rotations between any arbitrary tube-$i$ and tube-$1$ (outer tube) are related by the following relationship: $R_{i} = R_1R_{d_3}(\theta_{i})$, where $R_{d_3}(\theta_{i})$, is simply a rotation about $\bm{d_{3}}$ by an angle $\theta_{i}$, which represents the `twist angle' between the tube-$i$ and tube-1 (see Fig.~\ref{fig:dilation_twist}(b)). 
Note, for the purposes of this paper, $\theta_{1}$ is assumed to be zero, without loss of generality.

In addition to relative twisting, a pair of tendon-driven telescoping tubes also experiences relative axial dilation when actuated. The dilation factor of each tube, denoted as \( e_i \), can be expressed relative to the outermost tube as $\beta_i = \frac{e_i}{e_1}$. Note, by definition, $\beta_{1}$ is unity.

\subsection{Compatibility Constraints}
Since overlapping tubes share a common twist axis $\bm{d}_3$, their bending curvatures $\bm{u}_i|_{(\bm{d}_1,\bm{d}_2)}$ must be equal when expressed in a common coordinate frame. 
Consequently, this compatibility constraint reduces the number of independent state variables to \( \bm{u}_1 \) and the third components of the remaining tubes, \( \{ u_{i,d_3} \} \). By differentiating both sides of the relation \( R_i = R_1 R_{d_3}(\theta_i) \) with respect to arc length \( s \), and applying the kinematic equations from Eq.~(\ref{eq:shape_evol}), the relationship between \( \bm{u}_1 \) and \( \bm{u}_i \) can be derived as:  
\begin{align}\label{eq:u2_u1_relationship}
    \bm{u}_i(s) = R^T_{d_3}(\theta_i)\bm{u}_1(s) + \dot{\theta}_i(s)\mathbb{I}_3
\end{align}
It follows directly from \eqref{eq:u2_u1_relationship} that 
\begin{align*}
    \dot{\theta}_i(s) &= u_{i,d_3}(s) - u_{1,d_3}(s) \\
    \ddot{\theta}_i(s) &= \dot{u}_{i,d_3}(s) - \dot{u}_{1,d_3}(s)
\end{align*}
Differentiating Eq.~(\ref{eq:u2_u1_relationship}) yields:
\begin{dmath}\label{eq:der_equal_curvature}
\dot{\bm{u}}_i = R^T_{d_3}(\theta_i) \dot{\bm{u}}_1 + \frac{d R^T_{d_3}(\theta_i)}{d\theta_i} \dot{\theta}_i \bm{u}_1 + \ddot{\theta}_i \mathbb{I}_3 \\
= (R^T_{d_3}(\theta_i) - \mathbb{I}_{3}) \dot{\bm{u}}_1 + \dot{\theta}_i [\mathbb{I}_3]^T R^T_{d_3}(\theta_i) \bm{u}_1 + \mathbb{I}_3 \dot{u}_{i,d_3}
\end{dmath}
The term \( \ddot{\theta}_i \mathbb{I}_3 \) can be rewritten as:
\begin{align}
    \ddot{\theta}_i \mathbb{I}_3 = \mathbb{I}_3 \dot{u}_{i,d_3} - \mathbb{I}_{(3,3)}\dot{u}_{1,d_3}
\end{align}
where \( \mathbb{I}_{(3,3)} = \text{diag}(\mathbb{I}_3) \) is a diagonal matrix with the vector \( \mathbb{I}_3 \) on its diagonal.

Additionally, sharing a common centerline implies that the tangent vector \( \bm{t} \) is the same for all tubes. The relationship between the shear strains of tube \( i \) and the reference tube is given by:
\begin{align}\label{eq:shear_relation}
\bm{v}_1 = R_1^T(e_1\bm{t}), \quad
\bm{v}_i = R_i^T(e_i\bm{t}) = R_{d_3}^T(\theta_i)R_1^T(e_i\bm{t})
\end{align}
\[
\Rightarrow \bm{v}_i = \beta_i R^T_{d_3}(\theta_i)\bm{v}_1
\]
Differentiating Eq.~(\ref{eq:shear_relation}) with respect to arc length \( s \) gives:
\small
\begin{align}\label{eq:shear_dot_relation}
\dot{\bm{v}}_i = \dot{\beta}_i R^T_{d_3}(\theta_i)\bm{v}_1 + \beta_i [\mathbb{I}_3]^T R^T_{d_3}(\theta_i) \dot{\theta}_i \bm{v}_1 + \beta_i R^T_{d_3}(\theta_i) \dot{\bm{v}}_1
\end{align}
\normalsize
this is an important distinction compared to the derivation of \(\bm{v}_i\) and \(\dot{\bm{v}}_i\) by Chitalia et al.~\cite{chitalia2023modeling}, where the authors assumed that sharing a common centerline among tubes implies that the \(d_1\) and \(d_2\) components of \(\bm{v}_i\) and \(\bm{v}_1\) are related by a rotation about \(d_3\). They related the linear strains \(\bm{v}_i\) and \(\bm{v}_1\) analogously to the bending strains \(\bm{u}_i\) and \(\bm{u}_1\). However, this assumption is inaccurate.

Both \( \bm{u}_i \) and \( \bm{v}_i \), along with their derivatives, are expressed in terms of the reference tube's strain variables \( \bm{u}_1 \) and \( \bm{v}_1 \), and their respective derivatives.

\subsection{Equilibrium Equations}
For a special Cosserat rod with internal forces and moments indicated by vectors $\bm{n}$ and $\bm{m}$ respectively, and applied forces and moments per unit length given by $\bm{f}$ and $\bm{\tau}$ respectively, the equations for static equilibrium are indicated in \cite{dupont2009design} as:
\begin{align} \label{eq:equilibrium_eq_single_tube}
    \begin{bmatrix}
        \dot{\bm{n}}_i \\
        \dot{\bm{m}}_i
    \end{bmatrix}
    +
    \begin{bmatrix}
        [\bm{u}_i] & 0 \\
        [\bm{v}_i] & [\bm{u}_i]
    \end{bmatrix}  
    \begin{bmatrix}
        \bm{n}_i \\
        \bm{m}_i
    \end{bmatrix} + 
    \begin{bmatrix}
        \bm{f}_i \\
        \bm{\tau}_i
    \end{bmatrix}=0
\end{align}

\subsection{Constitutive Model}
For each tube, shear force can be related to shear strain and bending moment to curvature using a constitutive model. Assuming a linear relationship,
\begin{align}\label{eq:constitutive_model_single}
     \begin{bmatrix}
        \bm{n}_i \\
        \bm{m}_i
    \end{bmatrix} = 
    \begin{bmatrix}
        K_{se,i} & 0 \\
        0 & K_{bt,i}
    \end{bmatrix}     
    \begin{bmatrix}
        \bm{v}_i - \bm{v}^*_i \\
        \bm{u}_i - \bm{u}^*_i
    \end{bmatrix}     
\end{align}
and,
\small
\begin{eqnarray} \label{eq:shear_const}
    K_{se,i} & = & 
    \begin{bmatrix}
        GA_i & 0 & 0 \\
        0 & GA_i & 0 \\
        0 & 0 & EA_i
    \end{bmatrix} ,   \\ \label{eq:moment_const}
    K_{bt,i} & = & 
    \begin{bmatrix}
        EI_{xx,i} & 0 & 0 \\
        0 & EI_{yy,i} & 0 \\
        0 & 0 & GJ_{zz,i} 
    \end{bmatrix}    
\end{eqnarray}
\normalsize
where $\bm{u}^*_i$ and $\bm{v}^*_i$ represent the angular and linear strain vectors, respectively in the rest configuration, $G$ is the shear modulus, $E$ is the elastic modulus, $I_{xx}$ and $I_{yy}$ are the second moments of area about the local frame axes, while $J_{zz}$ is the corresponding polar moment of area. Ideally, $K_{se}$ and $K_{bt}$ are functions of the arc length $s$, but this assumption is dropped for the purposes of this paper similar to \cite{chitalia2023model}.

\subsection{Distributed Forces and Moments}
The distributed force \(\bm{f}_i\) and moment \(\bm{\tau}_i\) applied to each tube consist of three components: contributions from tendon actuation ($\bm{f}_{t}$, $\bm{\tau}_{t}$), interactions with adjacent tubes ($\bm{f}_{c}$, $\bm{\tau}_{c}$), and external loads ($\bm{f}_{e}$, $\bm{\tau}_{e}$). 
\begin{eqnarray} \label{eq:force_components}
    \bm{f}_i & = & \bm{f}_{t,i} + \bm{f}_{c,i} + \bm{f}_{e,i}, \\ 
    \bm{\tau}_i & = & \bm{\tau}_{t,i} + \bm{\tau}_{c,i} + \bm{\tau}_{e,i}
    \label{eq:moment_components}
\end{eqnarray}
Tendon position in the cross-sectional body frame can be expressed as $\bm{r(s)}_{j, i} = [x(s)_{j, i}, y(s)_{j, i}, 0]^T$($i_{th}$ tube and $j_{th}$ tendon). Following \cite{rucker2011statics}, the distributed force and moment components due to scalar tendon tension, $\lambda_{j, i}$ are given by

\begin{eqnarray} \label{eq:tendon_forces}
    \bm{f}_{t,j,i} = - \lambda_{j,i}\frac{[\dot{\bm{p}}_{j,i}^b]^2}{||\dot{\bm{p}}_{j,i}^b||^3}\ddot{\bm{p}}_{j,i}^b, \\
    \bm{\tau}_{t,j,i} =[\bm{r}_{j,i}]\bm{f}_{t,j,i} = - [\bm{r}_{j,i}]\lambda_{j,i}\frac{[\dot{\bm{p}}_{j,i}^b]^2}{||\dot{\bm{p}}_{j,i}^b||^3}\ddot{\bm{p}}_{j,i}^b
    \label{eq:tendon_moment}
\end{eqnarray}

where the derivatives of tendon position may be expressed (in the body frame) as follows:
\begin{align} \label{eq:tendon_position_derivatives}
    \dot{\bm{p}}_{j,i}^b = [\bm{u}_i]\bm{r}_{j,i} + \dot{\bm{r}}_{j,i} + \bm{v}_i, \nonumber \\
    \ddot{\bm{p}}_{j,i}^b = [\bm{u}_i]\dot{\bm{p}}_{j,i}^b - [\bm{r}_{j,i}]\dot{\bm{u}}_i + \dot{\bm{v}}_i + [\bm{u}_i]\dot{\bm{r}}_{j,i}+\ddot{\bm{r}}_{j,i} 
\end{align}
Assuming negligible friction, the tubes exert equal and opposite interaction forces along the common axis $\bm{d}_3$. As a result, the net internal forces and moments exchanged between the tubes in the $\bm{d_{1}}$ and $\bm{d_{2}}$ directions cancel out, and no interaction occurs along $\bm{d_{3}}$ axis. Additionally, in the absence of external loading, we assume $\bm{f}_{e,i} = \bm{\tau}_{e,i} = 0$. Consequently, the force and moment balance equations (\ref{eq:force_components}) and (\ref{eq:moment_components}) reduce to the following form in the $\bm{d}_{1}$ and $\bm{d}_{2}$ directions:
\begin{align} \label{eq:reduced_force_components}
    \bm{f}_i  =  \bm{f}_{t,i}, \quad
    \bm{\tau}_i  =  \bm{\tau}_{t,i} 
\end{align}

\subsection{Model Equations}\label{subsec:ModelEquations}
To sum all moments, they must be expressed in a common coordinate frame, for which the outermost tube's material frame is a natural choice. The transformation between any intermediate tube $i$ and the outermost frame is $R_i = R_1 R_{d_3}(\theta_i)$, allowing the moments to be expressed consistently and leading to the following formulation:
\begin{dmath} \label{eq:moment_sum_local1}
    \sum_{i = 1}^n R_{d_3}(\theta_i)\left(\dot{\bm{m}}_i + [\bm{u}_i]\bm{m}_i + [\bm{v}_i]\bm{n}_i + \bm{\tau}_c\right) = \mathbf{0} \big|_{\bm{d}_1, \bm{d}_2}
\end{dmath}

\noindent Substituting the constitutive model from Eq.~(\ref{eq:constitutive_model_single}) and distributed forces from Eq.~(\ref{eq:tendon_moment}) into the above equation, we get the following:
\small
\begin{dmath} \label{eq:moment_sum_local2}
    \sum_{i = 1}^n R_{d_3}(\theta_i)\left(K_{bt,i}(\dot{\bm{u}}_i - \dot{\bm{u}}^*_i)+ [\bm{u}_i]K_{bt,i}(\bm{u}_i - \bm{u}^*_i) +\\
    [\bm{v}_i]K_{se,i}(\bm{v}_i - \bm{v}^*_i) -\sum_{j = 1}^{m_i} [\bm{r}_{j,i}] \lambda_{j,i} \frac{[\dot{\bm{p}}_{j,i}^b]^2}{||\dot{\bm{p}}_{j,i}^b||^3} \ddot{\bm{p}}_{j,i}^b \right) = \mathbf{0} \big|_{\bm{d}_1, \bm{d}_2}
\end{dmath}
\normalsize
Substituting the expression for tendon position and its derivatives from Eq.~(\ref{eq:tendon_position_derivatives}), we get the following:
\small
\begin{dmath} \label{eq:moment_sum_local3}
    \sum_{i = 1}^n R_{d_3}(\theta_i)\Bigg(K_{bt,i}(\dot{\bm{u}}_i - \dot{\bm{u}}^*_i) + [\bm{u}_i]K_{bt,i}(\bm{u}_i - \bm{u}^*_i) + \\
    [\bm{v}_i]K_{se,i}(\bm{v}_i - \bm{v}^*_i) - \sum_{j = 1}^{m_i} \frac{\lambda_{j,i}}{||\dot{\bm{p}}^b_{j,i}||^3}[\bm{r}_{j,i}][\dot{\bm{p}}^b_{j,i}]^2 \\
    \left([\bm{u}_i]\dot{\bm{p}}^b_{j,i} - [\bm{r}_{j,i}]\dot{\bm{u}}_i + \dot{\bm{v}}_i + [\bm{u}_i]\dot{\bm{r}}_{j,i} + \ddot{\bm{r}}_{j,i}\right)\Bigg) = \mathbf{0} \big|_{\bm{d}_1, \bm{d}_2}
\end{dmath}
\normalsize
Substituting the compatibility constraints for curvature Eq.~(\ref{eq:der_equal_curvature}) and linear strain rates from Eq.~(\ref{eq:shear_dot_relation}) into Eq.~(\ref{eq:moment_sum_local3}), we get the following equation: 
\small
\begin{dmath} \label{eq:moment_sum_local4}
    \sum_{i = 1}^n R_{d_3}(\theta_i)\left(K_{bt,i}((R^T_{d_3}(\theta_i)-\mathbb{I}_{(3,3)})\dot{\bm{u}}_1+\\
    \dot{\theta}_i[\mathbb{I}_3]^TR^T_{d_3}(\theta_i)\bm{u}_1+\dot{u}_{d_3,i}\mathbb{I}_3-\dot{\bm{u}}^*_i)+[\bm{u}_i]K_{bt,i}(\bm{u}_i-\bm{u}^*_i)+\\
    [\bm{v}_i]K_{se,i}(\bm{v}_i-\bm{v}^*_i)-\sum_{j = 1}^{m_i}\frac{\lambda_{j,i}}{||\dot{\bm{p}}^b_{j,i}||^3}[\bm{r}_{j,i}][\dot{\bm{p}}^b_{j,i}]^2\\
    ([\bm{u}_i]\dot{\bm{p}}^b_{j,i}+[\bm{u}_i]\dot{\bm{r}}_{j,i}+ \ddot{\bm{r}}_{j,i}-[\bm{r}_{j,i}]((R^T_{d_3}(\theta_i)-\mathbb{I}_{(3,3)})\dot{\bm{u}}_1+\\
    \dot{\theta}_i[\mathbb{I}_3]^TR^T_{d_3}(\theta_i)\bm{u}_1+\dot{u}_{d_3,i}\mathbb{I}_3) + \dot{\beta}_iR^T_{d_3}(\theta_i)\bm{v}_1+\\
    \beta_i[\mathbb{I}_3]^TR^T_{d_3}(\theta_i)\dot{\theta}_i\bm{v}_1+\beta_iR^T_{d_3}(\theta_i)\dot{\bm{v}}_1)\right)=\mathbf{0}|_{\bm{d}_1,\bm{d}_2}
\end{dmath}
\normalsize
The above equation can be summarized as:
\small
\begin{dmath} \label{eq:moment_sum_local5}
G_u \cdot \dot{\bm{u}}_1 + G_v \cdot \dot{\bm{v}}_1 + G_{u_{d_3,i}} \mathbb{I}_3 \cdot \dot{u}_{d_3,i} + G_{\beta_i} \cdot \dot{\beta}_i = \bm{RHS}_1
\end{dmath}
\normalsize
Here, the matrices themselves are functions of the state
variables and are given by the following expressions:
\small
\begin{dmath} \label{eq:Gu}
    G_u|_{\bm{d}_1,\bm{d}_2} = K_{bt,1}+\sum_{j = 1}^{m_1}\frac{\lambda_{j,1}}{||\dot{\bm{p}}^b_{j,1}||^3}[\bm{r}_{j,1}][\dot{\bm{p}}^b_{j,1}]^2[\bm{r}_{j,1}]+\\
    \sum_{i = 2}^n R_{d_3}(\theta_i)K_{bt,i}(R^T_{d_3}(\theta_i)-\mathbb{I}_{(3,3)})+\\
    \sum_{j = 1}^{m_i}\frac{\lambda_{j,i}}{||\dot{\bm{p}}^b_{j,i}||^3}R_{d_3}(\theta_i)[\bm{r}_{j,i}][\dot{\bm{p}}^b_{j,i}]^2[\bm{r}_{j,i}](R^T_{d_3}(\theta_i)-\mathbb{I}_{(3,3)})
\end{dmath}
\vspace{1em}
\begin{dmath} \label{eq:Gv}
    G_v|_{\bm{d}_1,\bm{d}_2}=-\sum_{j = 1}^{m_i}\frac{\lambda_{j,1}}{||\dot{\bm{p}}^b_{j,1}||^3}[\bm{r}_{j,1}][\dot{\bm{p}}^b_{j,1}]^2\\
    -\sum_{i = 2}^n \sum_{j = 1}^{m_i}\frac{\lambda_{j,i}}{||\dot{\bm{p}}^b_{j,i}||^3}R_{d_3}(\theta_i)[\bm{r}_{j,i}][\dot{\bm{p}}^b_{j,i}]^2\beta_iR^T_{d_3}(\theta_i)
\end{dmath}
\vspace{1em}
\begin{dmath} \label{eq:Guz}
    G_{u_{d_3,i}}|_{\bm{d}_1,\bm{d}_2}=R_{d_3}(\theta_i)K_{bt,i}+\sum_{j = 1}^{m_i}\frac{\lambda_{j,i}}{||\dot{\bm{p}}^b_{j,i}||^3}R_{d_3}(\theta_i)[\bm{r}_{j,i}][\dot{\bm{p}}^b_{j,i}]^2[\bm{r}_{j,i}]
\end{dmath}
\vspace{1em}
\begin{dmath} \label{eq:Gbeta}
    G_{\beta_{i}}|_{\bm{d}_1,\bm{d}_2}=-\sum_{j = 1}^{m_i}\frac{\lambda_{j,i}}{||\dot{\bm{p}}^b_{j,i}||^3}R_{d_3}(\theta_i)[\bm{r}_{j,i}][\dot{\bm{p}}^b_{j,i}]^2R^T_{d_3}(\theta_i)\bm{v}_1
\end{dmath}
\vspace{1em}
\begin{dmath} \label{eq:RHS1}
    \bm{RHS}_1|_{\bm{d}_1,\bm{d}_2}= K_{bt,1}\dot{\bm{u}}^*_i-[\bm{u}_1]K_{bt,1}(\bm{u}_1-\bm{u}^*_1)-\\
    [\bm{v}_1]K_{se,1}(\bm{v}_1-\bm{v}^*_1)+\\
    \sum_{j = 1}^{m_1}\frac{\lambda_{j,1}}{||\dot{\bm{p}}^b_{j,1}||^3}[\bm{r}_{j,1}][\dot{\bm{p}}^b_{j,1}]^2([\bm{u}_1]\dot{\bm{p}}^b_{j,1}+[\bm{u}_1]\dot{\bm{r}}_{j,1}+ \ddot{\bm{r}}_{j,1})-\\
    \sum_{i = 2}^n R_{d_3}(\theta_i)K_{bt,i}(\dot{\theta}_i[\mathbb{I}_3]^TR^T_{d_3}(\theta_i)\bm{u}_1-\dot{\bm{u}}^*_i)-\\
    R_{d_3}(\theta_i)[\bm{u}_i]K_{bt,i}(\bm{u}_i-\bm{u}^*_i)-R_{d_3}(\theta_i)[\bm{v}_i]K_{se,i}(\bm{v}_i-\bm{v}^*_i)+\\
    \sum_{j = 1}^{m_i}\frac{\lambda_{j,i}}{||\dot{\bm{p}}^b_{j,i}||^3}R_{d_3}(\theta_i)[\bm{r}_{j,i}][\dot{\bm{p}}^b_{j,i}]^2([\bm{u}_i]\dot{\bm{p}}^b_{j,i}+[\bm{u}_i]\dot{\bm{r}}_{j,i}+ \ddot{\bm{r}}_{j,i})-\\
    \sum_{j = 1}^{m_i}\frac{\lambda_{j,i}}{||\dot{\bm{p}}^b_{j,i}||^3}R_{d_3}(\theta_i)[\bm{r}_{j,i}][\dot{\bm{p}}^b_{j,i}]^2[\bm{r}_{j,i}](\dot{\theta}_i[\mathbb{I}_3]^TR^T_{d_3}(\theta_i)\bm{u}_1)+\\
    \sum_{j = 1}^{m_i}\frac{\lambda_{j,i}}{||\dot{\bm{p}}^b_{j,i}||^3}R_{d_3}(\theta_i)[\bm{r}_{j,i}][\dot{\bm{p}}^b_{j,i}]^2\beta_i[\mathbb{I}_3]^TR^T_{d_3}(\theta_i)\dot{\theta}_i\bm{v}_1
\end{dmath}
\normalsize
The expression in Eq.~(\ref{eq:moment_sum_local5}) yields two equations that are linear in the derivatives of the state variables, \(\{\dot{\bm{u}}_1, \dot{\bm{v}}_1, \dot{u}_{d_3, i}, \dot{\beta}_i\}\). Additional independent scalar equations are obtained from the \( \bm{d}_3 \) component of the moment equilibrium for each tube. For each individual tube \( i \), this scalar equation takes the form:
\begin{dmath} \label{eq:moment_each_local1}
    \left\{\dot{\bm{m}}_i + [\bm{u}_i]\bm{m}_i + [\bm{v}_i]\bm{n}_i + \bm{\tau}_{c,i} = \mathbf{0} \right\}|_{\bm{d}_3}
\end{dmath}
A similar process is applied to the force equilibrium equations. A full expression of these is provided in the Appendix. The final form of the differential equation system is:

\begin{dmath} \label{eq:finall}
Ax =b
\end{dmath}
Where:
\begin{dmath} \label{eq:finall_A}
A=
\begin{bmatrix}
G_u & G_v & G_{u_{d_3,i}} \mathbb{I}_3 & G_{\beta_{i}} \\
{}_{i}H_u & {}_{i}H_v & {}_{i}H_{u_{d_3,i}}\mathbb{I}_3 &{}_{i}H_{\beta_{i}}\\
J_u & J_v & J_{u_{d_3,i}} \mathbb{I}_3 & J_{\beta_{i}} \\
{}_{i}K_u & {}_{i}K_v & {}_{i}K_{u_{d_3,i}}\mathbb{I}_3 &{}_{i}K_{\beta_{i}}\\
\end{bmatrix}\\
\end{dmath}

\begin{dmath} \label{eq:finall_x}
x=
\begin{bmatrix} \dot{u}_1 \\ \dot{v}_1 \\ \dot{u}_{d_3,i} \\ \dot{\beta}_{i} \end{bmatrix}
\end{dmath}

\begin{dmath} \label{eq:finall_b}
b=
 \begin{bmatrix} 
 RHS_1 \\  
 RHS_{2,i} \\ 
 RHS_3 \\  
 RHS_{4,i} 
 \end{bmatrix} 
 \end{dmath}
\normalsize
The dimensions of \( A \), \( x \), and \( b \) depend on the number of concentric tubes (\(n\)); increasing the number of tubes adds additional state variables and corresponding equations.

\subsection{Numerical solution}
Eq.~(\ref{eq:finall}) is numerically solved using a least-squares approach, which provides a minimum-norm solution to $Ax - b = 0$. For solving this differential equation, the initial conditions of the state variable are required. Since these values are not given, the shooting method\cite{osborne1969shooting} is deployed to guess the initial condition. By having $\lambda_{j, i}$ and boundary conditions at the base consist of the orientation and position $R(0)$,  $\bm{\dot{p}_{c}(0)}$ we attempt to solve the equation. Tendon termination at $s = l_i$, exerts a point
force and moment given by: 
\begin{eqnarray} \label{eq:boundry_tendon_forces}
    \bm{f}_{j,i}(l_i) = - \lambda_{j,i}\frac{\dot{\bm{p}}_{j,i}^b(l_i)}{||\dot{\bm{p}}_{j,i}^b(l_i)||}\\
    \bm{\tau}_{j,i}(l_i) = - [\bm{r}_{j,i}]\lambda_{j,i}\frac{\dot{\bm{p}}_{j,i}^b(l_i)}{||\dot{\bm{p}}_{j,i}^b(l_i)||}
    \label{eq:boundry_tendon_moment}
\end{eqnarray}
The boundary condition at the endpoint is defined as follows:
\begin{align}\label{eq:boundary_n}
 \bm{n}_i(l^-_i) = \bm{n}_i(l^+_i) + \bm{f}_{i}(l_i)
\end{align}
\begin{align}\label{eq:boundary_m}
 \bm{m}_i(l^-_i) = \bm{m}_i(l^+_i) + \bm{\tau}_{i}(l_i)
\end{align}
In these equations, $l^{\mp}_i$ denotes the arc-length values just before and after $l_i$. Any assumption that satisfies Eq.~(\ref{eq:boundary_n}) and (\ref{eq:boundary_m}) at the endpoint is considered a valid initial guess for Eq.~(\ref{eq:finall}), which can subsequently be solved numerically.

\section{Experimental Model Validation}\label{sec:Experimental_Model_Validation}
\begin{figure}[t]
\centering\includegraphics[width=\columnwidth]{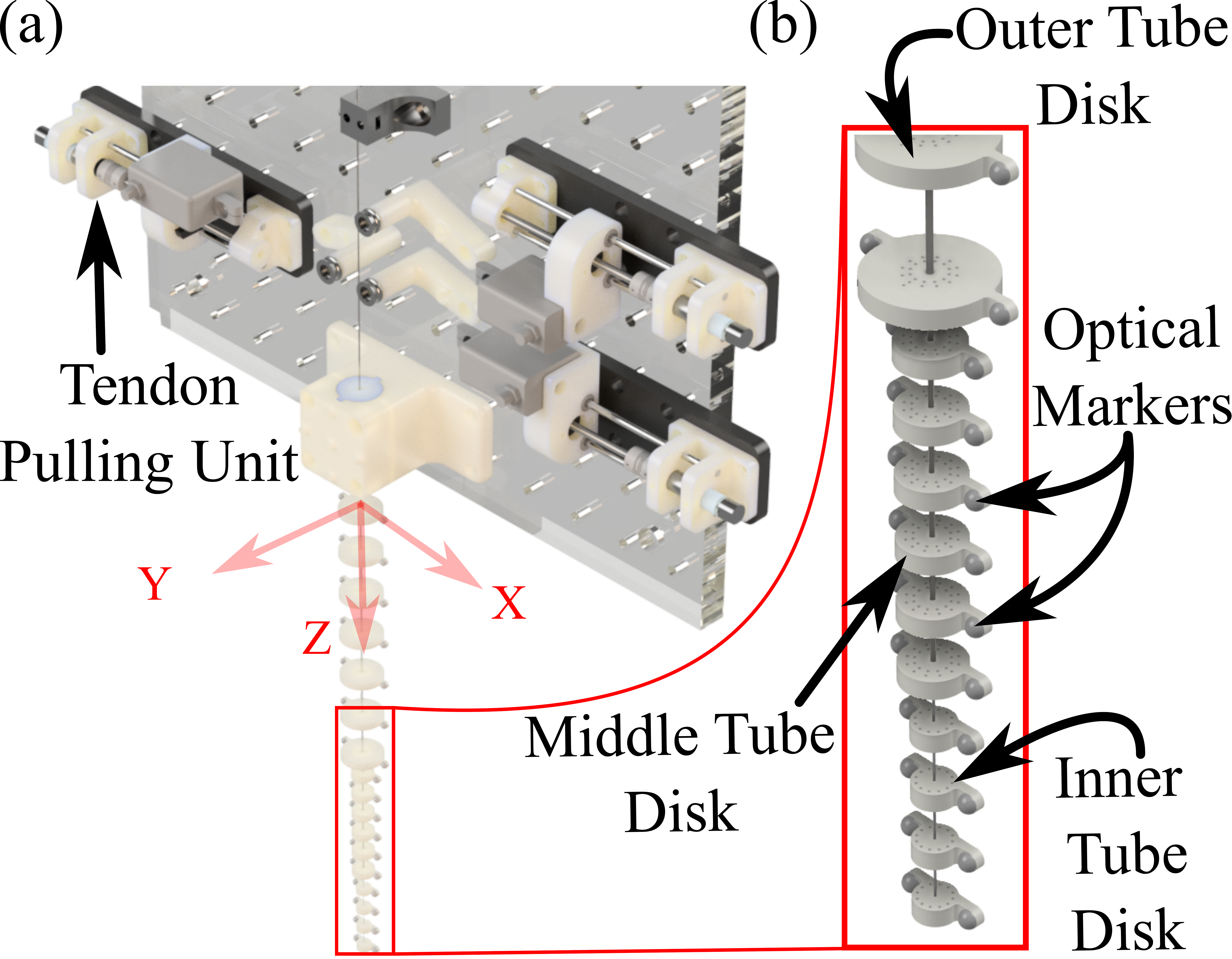}
\caption{CAD model of the experimental setup used for model validation. (a) Overall system configuration, including the tendon pulling unit and tube assembly. (b) Close-up view of the disks mounted on each tube (outer, middle, and inner), each featuring optical markers on their sides for shape tracking.}
\label{fig:ExperimentalSetup}
\vspace{-5 mm}
\end{figure}

This section presents a comprehensive experimental evaluation of the proposed model across multiple robot platforms and configurations. The experiments include a developed two-tube and three-tube configuration with varied tendon routing strategies, as well as evaluations on different robots designed for use in the medical field. These include a pre-curved concentric tube robot (CTR)~\cite{dupont2009design} and two tendon-actuated concentric tube robots: TACTER~\cite{yamamoto2025tacter} and ExoNav II~\cite{moradkhani2025exonav}. These platforms span a range of mechanical structures and tendon routing conditions. The model is evaluated in each case by comparing the predicted robot shapes to shapes obtained from experimental measurements, in order to assess modeling accuracy and generality.
Figure~\ref{fig:ExperimentalSetup}(a) illustrates the common setup for the two- and three-tube experiments. At the back end, three tendon-pulling units were installed, each actuating a single tendon. Each unit features a DC motor ($\phi$8~mm, Maxon Metal Brushes, $0.5$ Watt) driving a lead screw to advance or retract a load cell (MDB-$5$, $5$lb capacity, Transducer Techniques). The load cell’s linear motion applies or releases tension in the attached nitinol tendon. The system is modular and can be configured for both two-tube and three-tube experiments.
Each tube is guided by multiple disks, which include two flat circular surfaces on their sides to accommodate optical markers (see Fig.~\ref{fig:ExperimentalSetup}(b)). These markers were tracked using four Vero v2.2 motion capture cameras (Vicon Motion Systems Ltd., United Kingdom). The midpoint of each pair of markers determines the centerline of the robot.

\subsection{Two-Tube Configuration}
\begin{figure*}[]
\centering\includegraphics[scale=0.37, angle = 0]{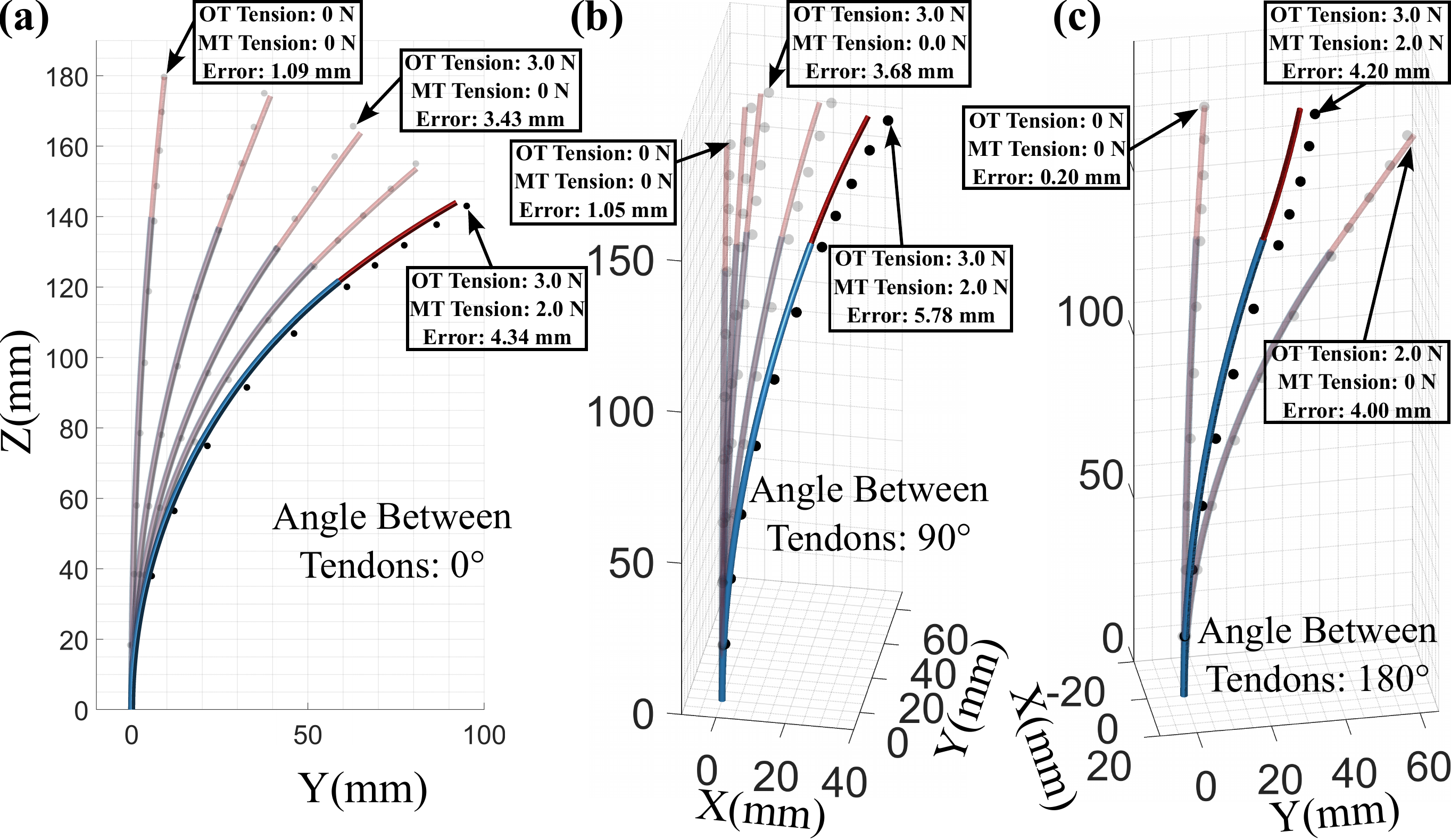}
\caption{Comparison between model-predicted robot shape and experimental results for the two-tube configuration. Black circles represent the midpoints of tracked marker pairs. The blue tube corresponds to the outer tube and the red tube to the inner tube. (a) Tendons routed in the same direction ($0\degree$ angle between tendons path), causing both tubes to bend along the global Y-axis, (b) tendon paths are oriented at a $90\degree$ angle to each other, resulting in orthogonal bending, (c) tendons are routed $180\degree$ apart, producing opposing bending directions.}
\label{fig:Results_TwoTube}
\end{figure*}
An initial experiment was conducted using a two-tube setup. One tendon was terminated at the end of each tube ($n=2$ and $m_i=1, \forall i$). The outer tube (tube 1) was a nitinol tube with length $140$~mm, elastic modulus $E_1=45~\text{GPa}$, and shear modulus $G_1 = 16.91~\text{GPa}$. The inner tube (tube 2) was a stainless steel tube with length $180$~mm, $E_2=215~\text{GPa}$, and $G_2 = 80~\text{GPa}$.
The reference global frame axes are shown in Figure~\ref{fig:ExperimentalSetup}(a). In the first trial, both tendons were routed along the local $\bm{d}_2$ direction, resulting in both tubes bending in the same global direction (Y-axis), as shown in Figure~\ref{fig:Results_TwoTube}(a). The robot at rest is first shown, followed by the actuation of the outer tube with a tendon tension of $3$~N. At this stage, the inner tube remains at its pre-curvature, as it has not yet been actuated. When the tendon tension at the distal ends of the outer and inner tubes reached $3$~N and $2$~N, respectively, the tip position error was measured to be $4.34$~mm, corresponding to approximately $2\%$ of the total robot length. The total robot tip displacement relative to the rest configuration was $90.14$~mm.
In the next trial, the tendons were routed at a $90\degree$ angle apart. Initially, actuation of the outer tube resulted in bending in the Y direction; subsequent actuation of the inner tube introduced additional bending in the X direction (see Fig.~\ref{fig:Results_TwoTube}(b)). With outer and inner tube tensions at $3$~N and $2$~N, respectively, the tip error was $5.78$~mm, approximately $3\%$ of the total robot length.
In the final trial (see Fig.~\ref{fig:Results_TwoTube}(c)), the tendons were routed at a $180\degree$ angle apart, causing the two tubes to bend in opposing directions. With outer and inner tube tensions at $3$~N and $2$~N, respectively, the tip error was $4.20$~mm, approximately $2\%$ of the total robot length.

\vspace{1 em}
\normalsize
\subsubsection{Helical Tendon Routing}
\begin{figure}[]
\centering\includegraphics[width=\columnwidth]{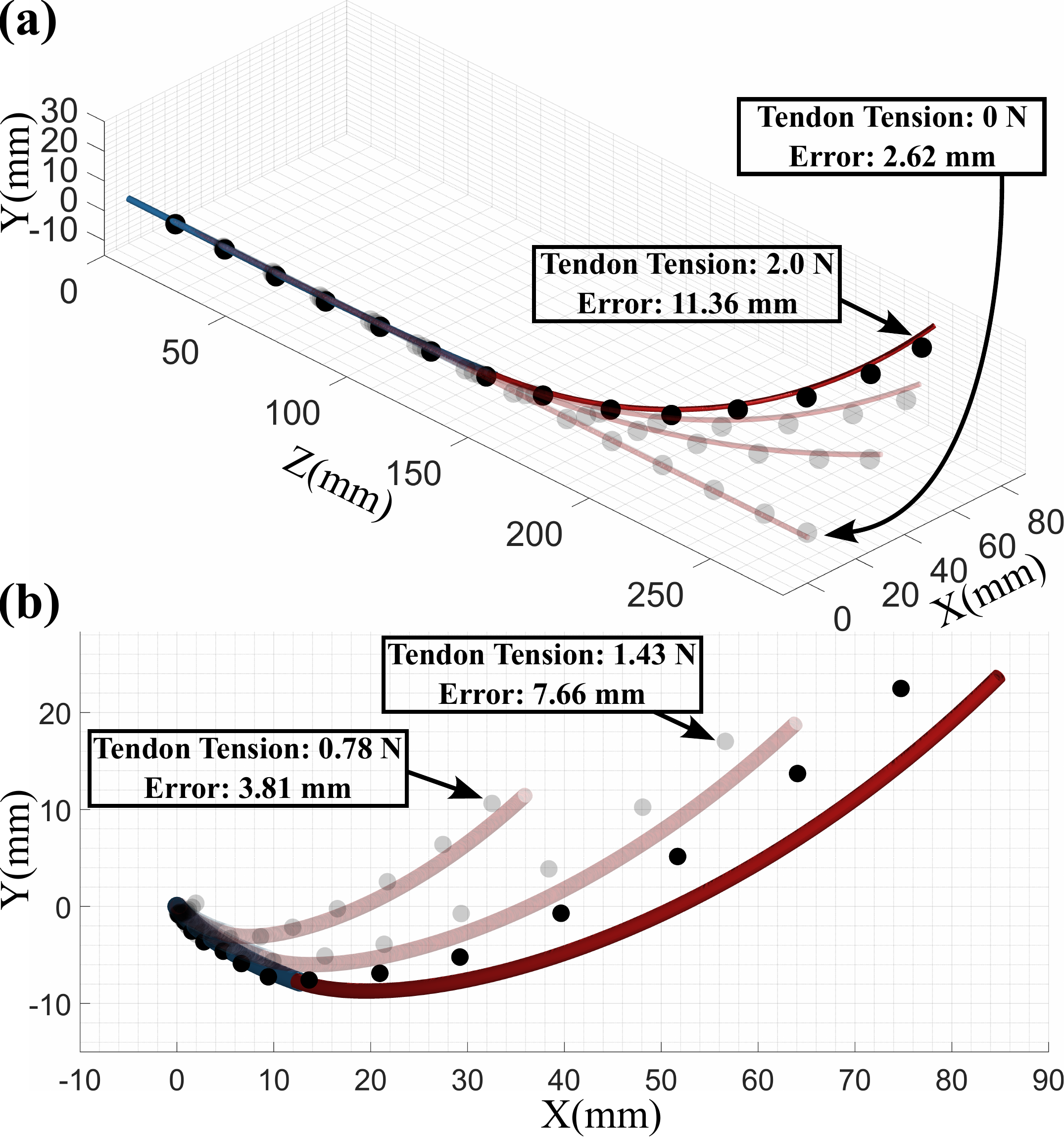}
\caption{Comparison between model-predicted and experimentally observed shape for a two-tube configuration with helical tendon routing. (a) Isometric view, (b) top view.}
\label{fig:Results_TwoTube_Helical}
\end{figure}
To evaluate the model under non-static tendon routing conditions, where \( \dot{\bm{r}} \) and \( \ddot{\bm{r}} \) are nonzero, a separate experiment was conducted. Since stainless steel tends to retain helical pre-curvatures due to its lack of super-elasticity, two nitinol tubes were used in this trial ($n=2$).
The outermost tube from the previous experiments, with length $140$~mm, was reassigned as the inner tube, and a new nitinol tube with length $280$~mm (OD = $1.35$~mm, ID = $1.07$~mm, $E=65~\text{GPa}$, $G=24.4~\text{GPa}$) was introduced as the outer tube.
A new set of disks was designed with $10\degree$ angle spacing between consecutive holes, allowing for a smoother helical routing along the robot.  
Seven disks were attached to each tube with $20$~mm spacing, resulting in 14 disks distributed along the robot and one additional disk at the base. A single tendon was routed helically ($m_1=0,~m_2 = 1$): starting at the tip in the local \( \bm{d}_2 \) direction, then routed through each subsequent disk with a $10\degree$ angular increment. The tendon routing is described by:
\small
\begin{equation*}
r_{1,1} = \begin{bmatrix}
0\\
0\\
0
\end{bmatrix}
, r_{1,2} = 6.5\times10^{-3}\cdot\begin{bmatrix}
\cos\left(\frac{2\pi s}{0.720} + \frac{31\pi}{18}\right) \\
\sin\left(\frac{2\pi s}{0.720} + \frac{31\pi}{18}\right) \\
0
\end{bmatrix}
\end{equation*}
\normalsize
The tendon was actuated to a tension of $2$~N. The model successfully predicted the robot's shape under these conditions, with a tip error of $11.36$~mm, corresponding to $3.93\%$ of the robot's total length (see Fig.~\ref{fig:Results_TwoTube_Helical}).

\vspace{1em}
\subsubsection{Hysteresis Effects}
\begin{figure}[ht]
\centering\includegraphics[width=\columnwidth]{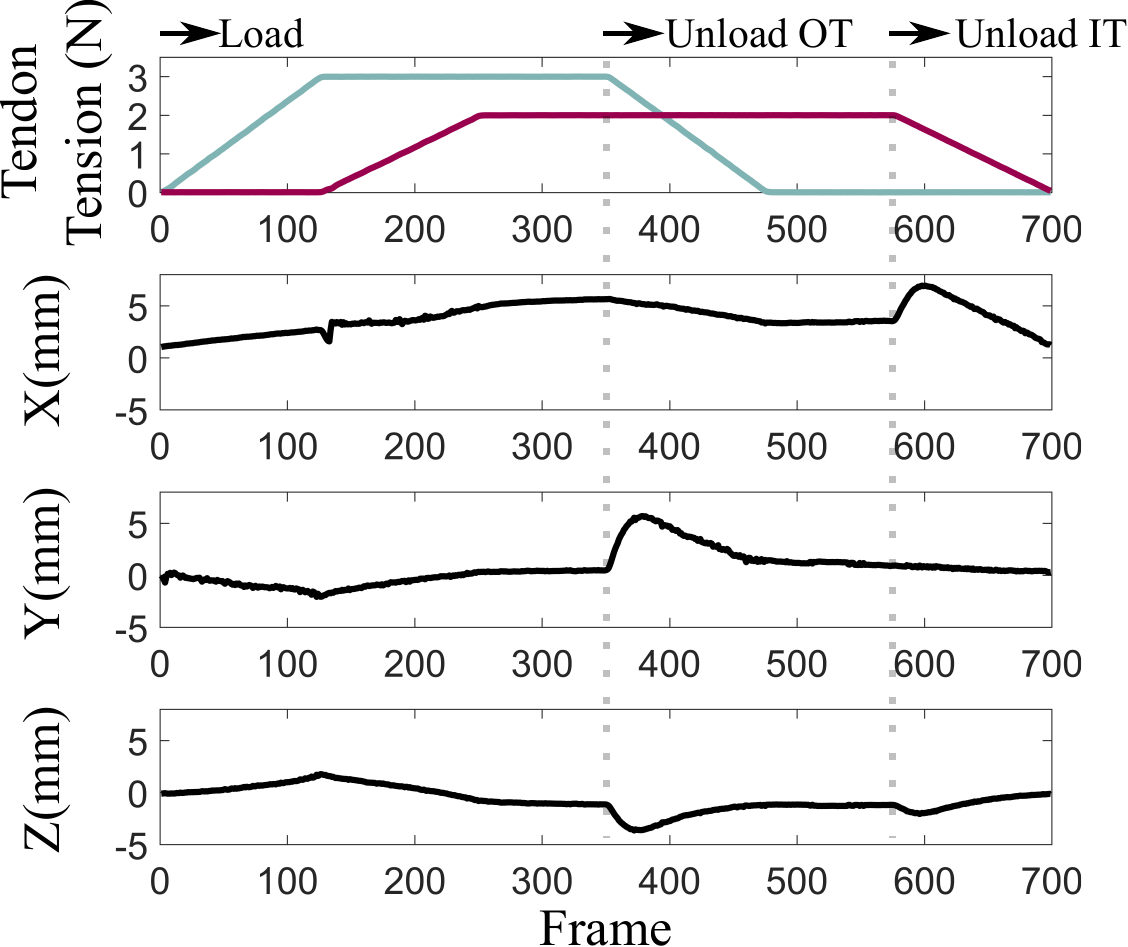}
\caption{Tip position error during loading and unloading of tendon tensions in a two-tube robot. The error increases smoothly during loading and rises more significantly during unloading due to hysteresis. Distinct error trends in the X and Y directions reflect the directional influence of each actuated tendon.}
\label{fig:Load_vs_Unload}
\vspace{-3 mm}
\end{figure}
Figure~\ref{fig:Load_vs_Unload} illustrates the tip error behavior of the two-tube configuration ($n=2$ and $m_i=1, \forall i$) during both the loading and unloading phases of tendon actuation. OT and IT refer to the outer and inner tube tendons, respectively. During loading, as tendon tensions increase, the tip error grows smoothly due to increasing overall deformation. At lower deformations, the robot behaves more linearly, leading to smaller errors, while larger deformations introduce more nonlinear effects. In the unloading phase, the error increases more significantly due to mechanical hysteresis. When the outer tube tendon (OT), routed along the $\bm{d}_2$ direction and primarily responsible for bending the tubes in Y, is released, a sharp increase in Y-direction error is observed. Releasing the inner tube tendon (IT), which governs bending in the X direction, causes a corresponding rise in X-direction error.

\subsection{Three-Tube Configuration}
\begin{figure*}[]
\centering\includegraphics[scale=0.4, angle = 0]{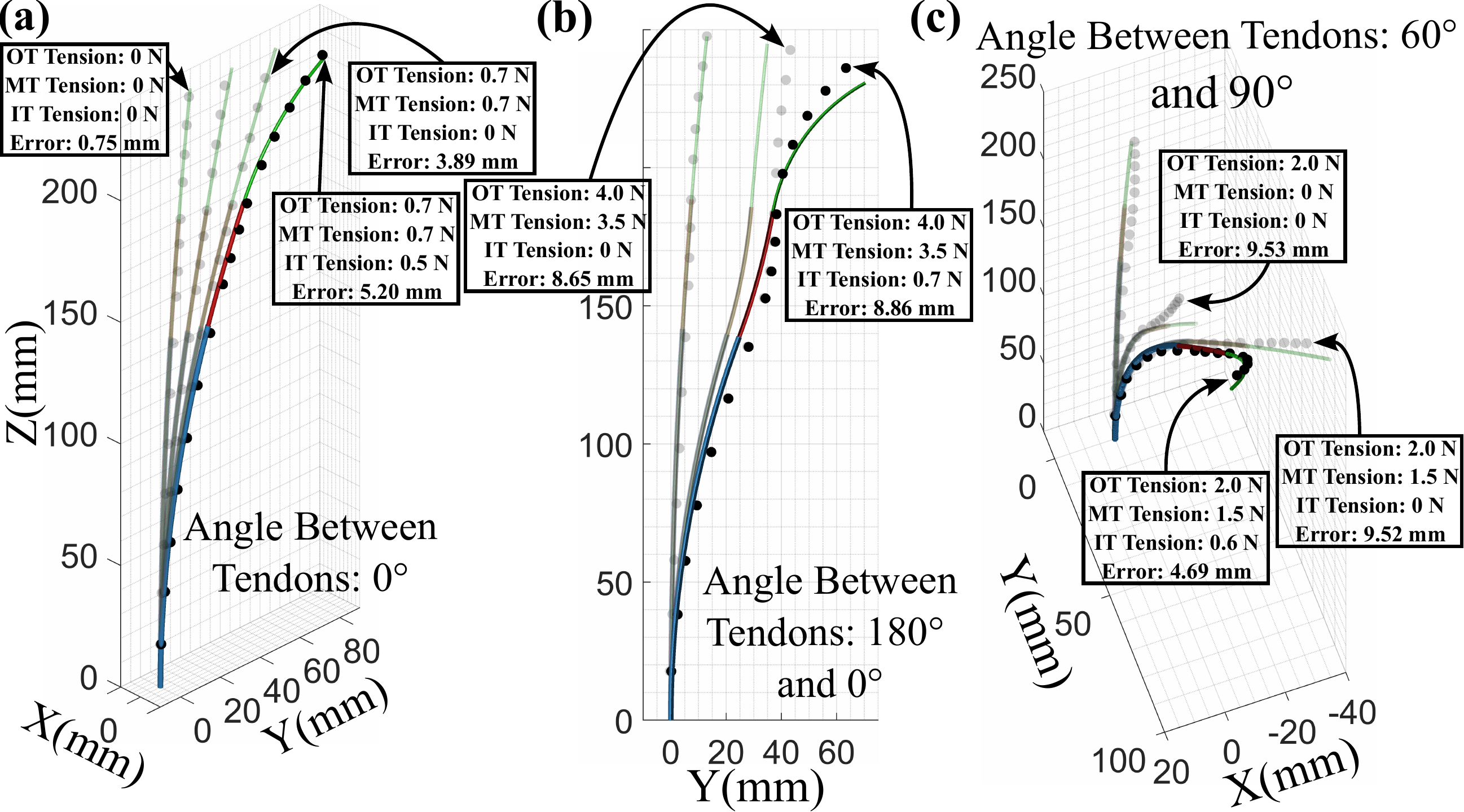}
\caption{Comparison between model-predicted and experimentally observed shapes for the three-tube configuration. Black circles indicate the midpoints of tracked marker pairs. The blue, red, and green tubes represent the outer, middle, and inner tubes, respectively. All tendon angles are defined with respect to the positive X-axis. (a) All tendons are routed in the same direction ($90\degree$), resulting in uniform bending. (b) Tendons routed at $90\degree$, $270\degree$, and $90\degree$ for the outer, middle, and inner tubes, respectively. (c) Tendons routed at $90\degree$, $150\degree$, and $0\degree$, creating complex and opposing curvatures.}
\label{fig:Results_ThreeTube}
\end{figure*}

Following the two-tube experiments, three-tube trials were conducted ($n=3$ and $m_i=1, \forall i$) by adding the innermost tube to the assembly. The mechanical properties of the outermost and middle tubes remained the same as in the two-tube trials, with the length of the middle tube updated to $187$~mm. The innermost tube (tube 3) was a stainless steel rod with a length of $246.22$~mm, an elastic modulus of $E_3 = 255~\text{GPa}$, and a shear modulus of $G_3 = 98~\text{GPa}$.
These values were determined experimentally, as approximately $14.5$~mm of the $59.22$~mm rod extended from the middle tube was covered by disks. This coverage significantly affects the bending stiffness, making nominal material properties insufficiently accurate for modeling purposes.

In the first trial, similar to the two-tube setup, all tendons were routed in the same direction (local $\bm{d}_2$), resulting in bending along the global Y-axis (see Fig.~\ref{fig:Results_ThreeTube}(a)). In subsequent trials, the tendon routing for the middle and inner tubes was modified, as illustrated in Fig.~\ref{fig:Results_ThreeTube}(b) and (c). The specific routing configurations and corresponding shapes are shown in the figure, with all tip position errors remaining within $4\%$ of the total robot length.

\subsubsection{Evaluation of Tendon Assignment Methods} \label{sec:TendonAssignment}

\begin{figure}[]
\centering\includegraphics[width=\columnwidth]{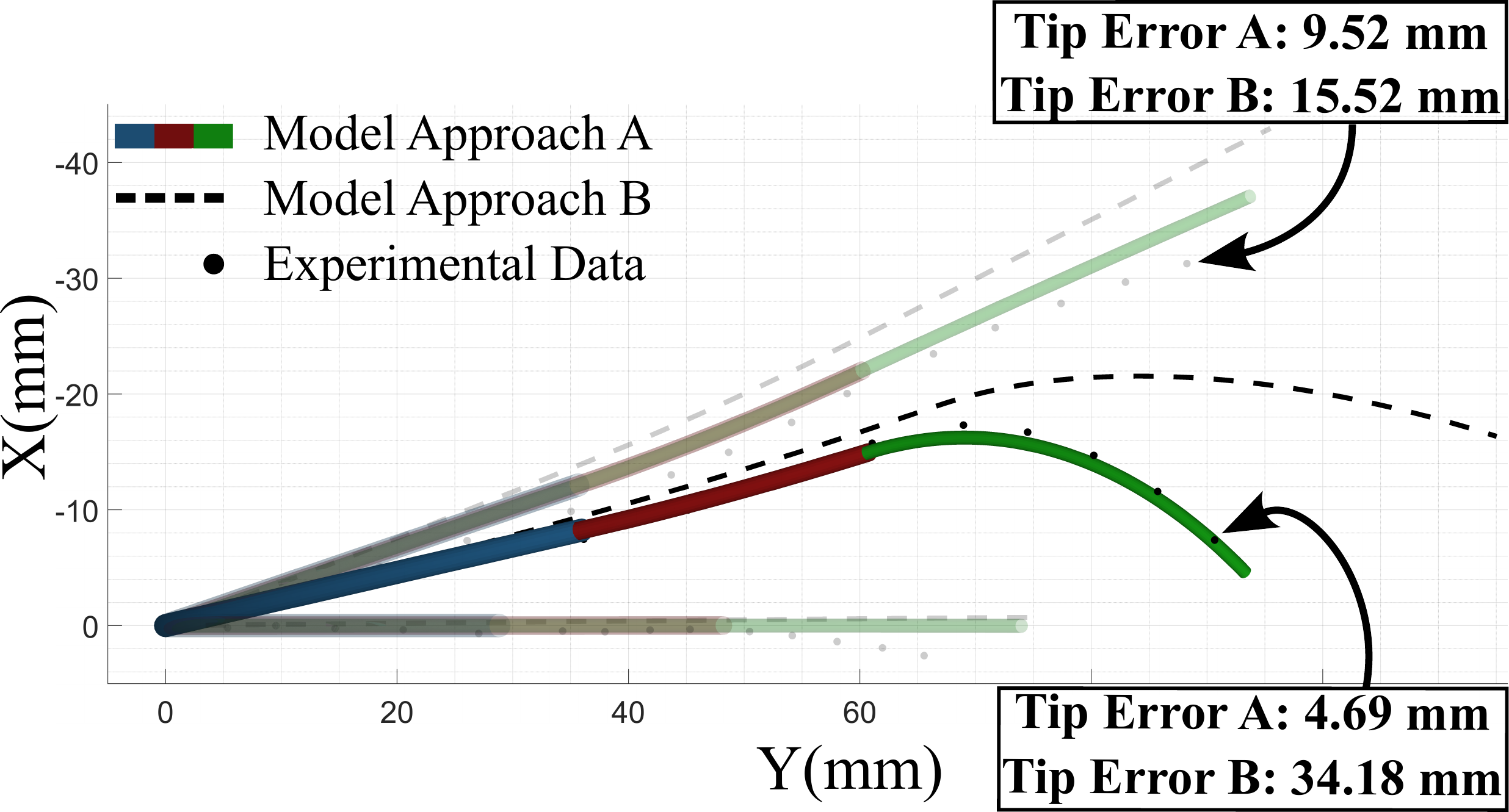}
\caption{Top view of a three-tube experiment comparing two modeling approaches: (A) assigning all tendons to the outermost tube, and (B) assigning each tendon to the tube where it terminates. When only the outermost tendon is actuated, both approaches yield similar results. However, when all tendons are actuated, Approach A shows significantly better agreement with the experimental data.}
\label{fig:Old_Vs_New_Approach}
\vspace{-5 mm}
\end{figure}
While most TACTRs assign tendons to specific tubes—routing each tendon within its corresponding tube—in our experimental setup, all tendons apply force and moment to the outermost tube within each section. This means that, during modeling, all tendons in a given section of concentric tubes are assigned to the outermost tube, while the inner tubes are not associated with any tendons. After solving the model for a given section, the boundary condition at the distal end is applied using only the tendon that is terminated at the tip of the outer tube, as it is the sole contributor to the boundary forces and moments.
Figure~\ref{fig:Old_Vs_New_Approach} compares two modeling approaches: (A) assigning all tendons to the outermost tube, and (B) assigning each tendon to the tube where it terminates. The experimental result is included for reference. When all tendons were actuated, Approach A demonstrated significantly better agreement with experimental data, yielding a tip error of $4.69$~mm compared to $34.28$~mm for Approach B. This improved accuracy is not only supported by experimental results, but is also consistent with the physical setup: in our design, all tendons pass through and mechanically interact with the disks of the outermost tube. As a result, the outermost tube bears the full tendon-induced forces and moments, making it the logical and physically accurate choice for modeling tendon assignment in our system. 

Although the proposed model is general and applicable to any \(n\)-tube configuration, incorporating tendon routing specifics is essential to ensure accurate modeling for a given robot.

\subsection{Concentric Tube Robot}
The proposed model is general and capable of simulating a wide range of concentric tube architectures, including concentric tube robots (CTRs) as a special case of tendon-actuated concentric tube robots (TACTRs). To demonstrate this, we validated the model on a CTR originally presented by Dupont et al.~\cite{dupont2009design}, which consists of two pre-curved tubes without any tendon actuation ($n=2$ and $m_i=0, \forall i$). In this case, the model can be applied without modification, highlighting its ability to accurately capture the behavior of concentric tube systems regardless of the actuation method.
\begin{figure}[t]
\centering\includegraphics[width=\columnwidth]{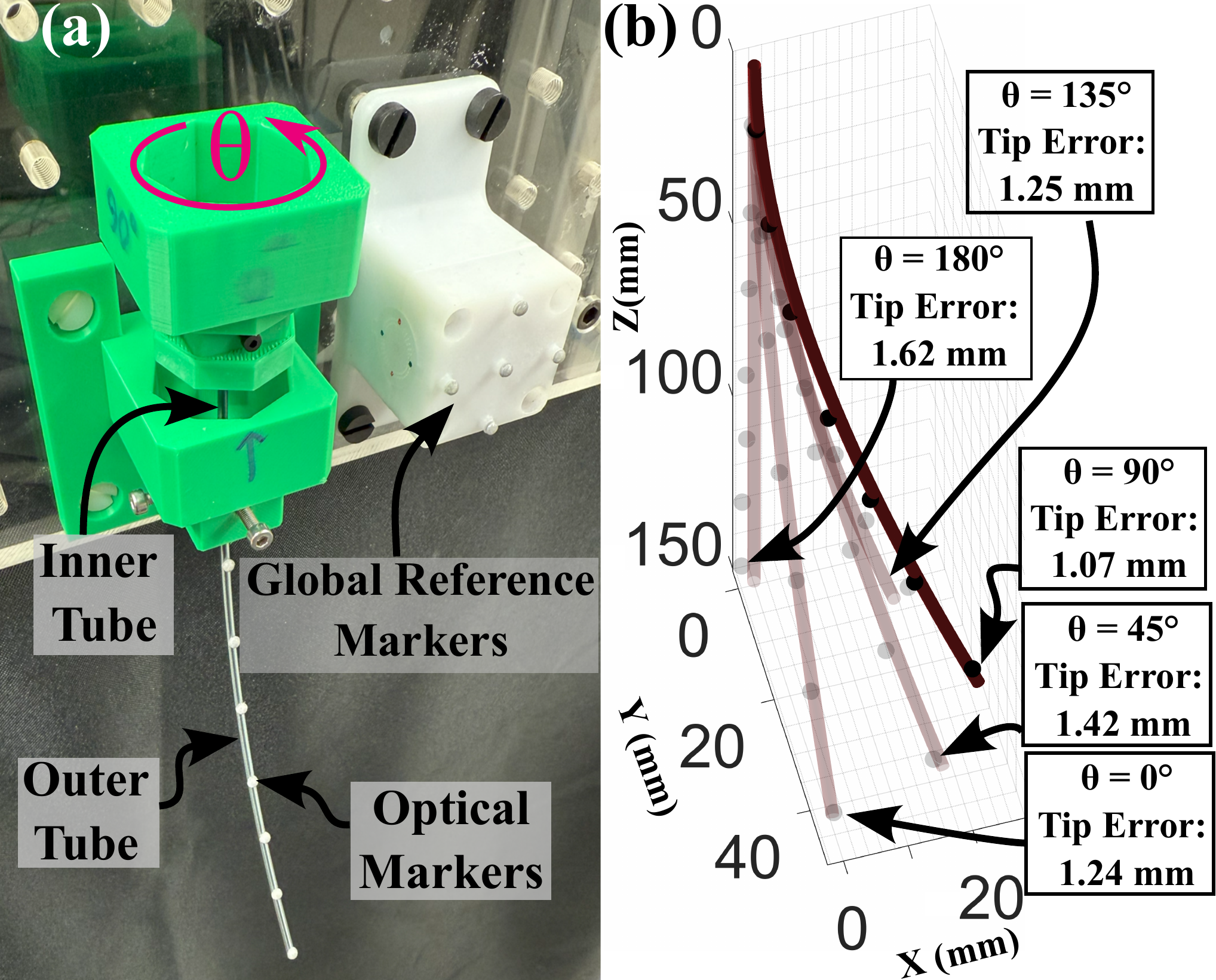}
\caption{CTR composed of two pre-curved tubes without tendon actuation~\cite{dupont2009design}. (a) Setup showing optical markers used to track the robot shape. The inner tube is mounted to an octagonal base, enabling five discrete rotational alignments relative to the outer tube. (b) Comparison between model-predicted and experimentally observed shapes for the five configurations. The model shows strong agreement, with tip errors ranging from $1.07$~mm to $1.62$~mm.}
\label{fig:CTR_Validation}
\vspace{-5 mm}
\end{figure}
The experimental setup is shown in Figure~\ref{fig:CTR_Validation}(a), where the inner tube is mounted to an octagonal male connector, which inserts into an octagonal female socket fixed to the outer tube. This configuration allows eight discrete rotational alignments between the two tubes. For validation, we selected five representative configurations based on discrete angular rotations between the inner and outer tubes. Initially, the two tubes were aligned, corresponding to a twist angle $\theta = 0\degree$. Subsequent configurations increased this twist angle in 45-degree increments, resulting from the octagonal interface design.
The mechanical properties of the tubes were designed to replicate those reported in~\cite{dupont2009design}, and any minor discrepancies—(particularly in the stiffness ratios and tube curvatures) were resolved by calibrating the tube parameters using the procedure described in the work. Specifically, we determined a stiffness ratio of $1.01$ (compared to $1.28$ reported in ~\cite{dupont2009design}) and an outer tube curvature of $219$~mm (measured using the experimental setup shown in Figure~\ref{fig:CTR_Validation}(a)). Figure~\ref{fig:CTR_Validation}(b) shows the model-predicted and experimental robot shapes across five configurations. The model accurately captures the robot’s shape with a maximum tip error of $1.62$~mm, corresponding to approximately $1.08\%$ of the total robot length. This validation demonstrates the applicability of our model to non-tendon-actuated concentric tube robots.

\subsection{Tendon-actuated Concentric Tube Endonasal Robot (TACTER)}
\begin{figure}[t]
\centering\includegraphics[width=\columnwidth]{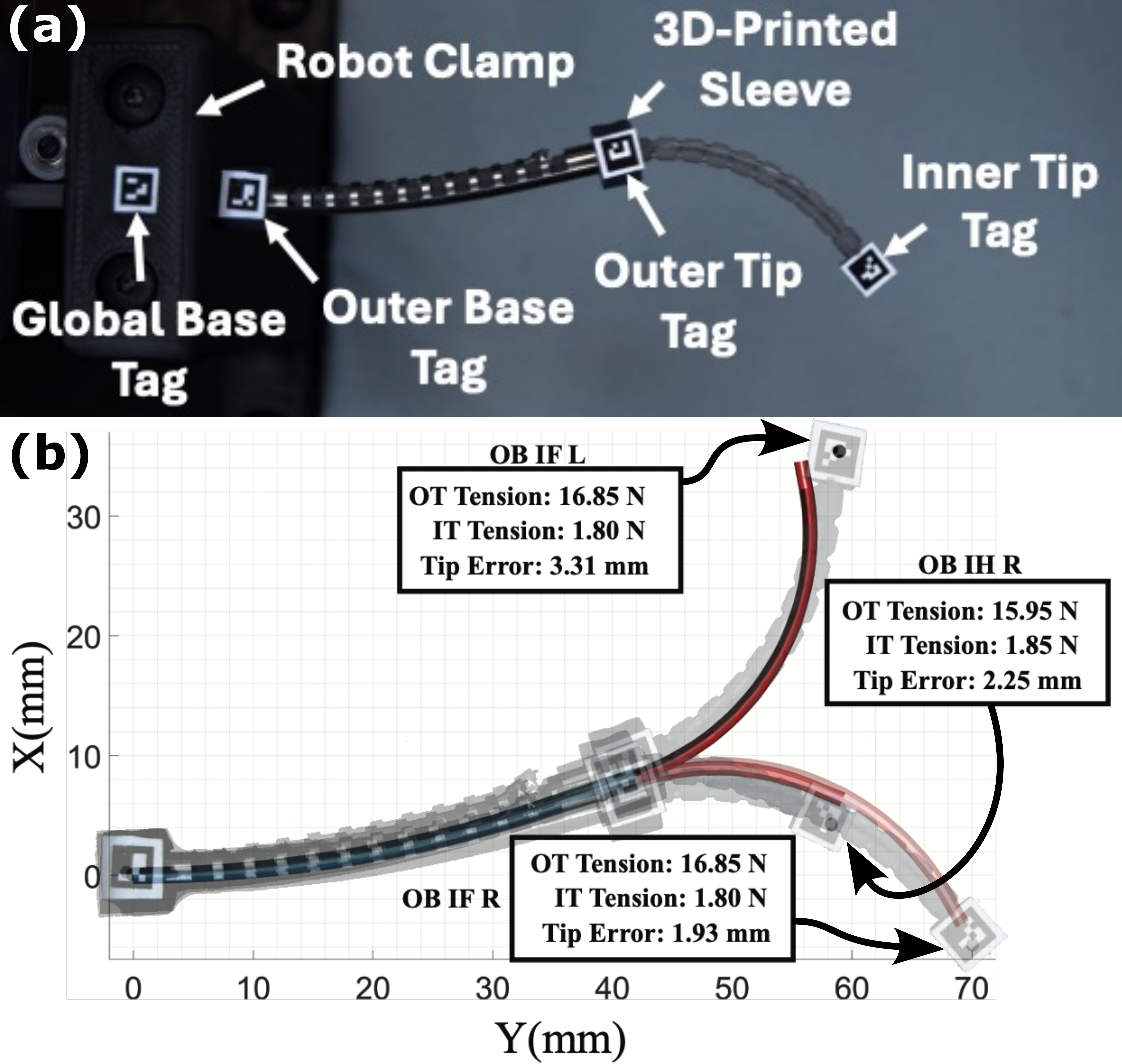}
\caption{TACTER, a tendon-actuated concentric tube robot developed for endonasal procedures~\cite{yamamoto2025tacter}. (a) Setup showing the two-tube robot with ArUco markers attached for vision-based tracking. (b) Comparison of model-predicted and experimentally observed robot shapes, with the actual robot shape overlaid on the predicted configuration. Outer tube tendon tensions (OT), inner tube tendon tensions (IT), and corresponding tip errors are indicated for each case.}
\label{fig:TACTER_Validation}
\vspace{-5 mm}
\end{figure}
In this section, we evaluate the proposed model using the TACTER robot, a tendon-actuated concentric tube system designed for endonasal procedures~\cite{yamamoto2025tacter}. The TACTER consists of two concentric robots ($n=2$, $m_1=1$ and $m_2=2$): an outer unidirectional asymmetric notched (UAN) nickel-titanium tube and an inner bidirectional 3D-printed robot with a nickel-titanium bending member (see Fig.~\ref{fig:TACTER_Validation}(a)). The inner robot can axially translate within the outer robot, enabling follow-the-leader motion through complex anatomical pathways. Detailed geometric and mechanical parameters are provided in~\cite{yamamoto2025tacter}.
We tested three configurations combining outer robot bending (OB), partial or full translation of the inner robot (IH/IF), and directional tendon actuation (left/right: L/R).
A vision-based approach is used to track the position of key TACTER features using 4 ArUco tags - global base, outer base, outer tip, and inner tip (see Fig.~\ref{fig:TACTER_Validation}(a)). 3D-printed sleeves are used to rigidly fix the ArUco tags on the outer robot, and a 3D-printed end-effector attachment that fits inside the inner robot lumen holds the inner tip ArUco tag. The 1800 U-2050C camera (from Allied Vision Technologies GmbH, Stadtroda, Germany) 
is used to capture images at each pose, then the position of each ArUco tag is obtained and normalized with respect to the "base" tag. The translational offsets due to the ArUco tag attachment methods have been applied in data post-processing.

Figure~\ref{fig:TACTER_Validation}(b) shows the model prediction results for all three experiments, with tendon tensions indicated in each case. The outer tube, being significantly stiffer than the inner, remains nearly unchanged across trials; this is why its predicted shapes overlap and appear indistinguishable. This mechanical separation is advantageous for path planning in TACTRs.
Our model accurately predicted the robot shape across all trials, with the maximum tip error reaching only $3.31$~mm, corresponding to $4.37\%$ of the robot's length.
Notably, in TACTER, each tendon is routed through its respective tube, unlike our experimental setup where tendons are assigned to the outermost tube (see Section~\ref{sec:TendonAssignment}).

\subsection{ExoNav II Robot}

\begin{figure}[t]
\centering\includegraphics[width=\columnwidth]{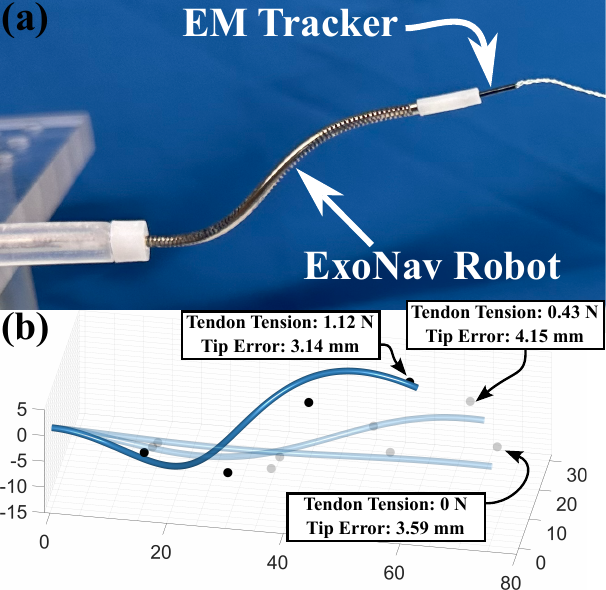}
\caption{ExoNav II, a tendon-actuated concentric tube robot designed for spinal cord stimulation~\cite{moradkhani2025exonav}. (a) Setup showing the robot with electromagnetic tracker (EM) used to capture the robot shape. (b) Comparison of model-predicted and experimentally observed robot shapes at multiple tendon tensions. Tip errors are shown for each configuration.}
\label{fig:ExoNav_v2}
\vspace{-5 mm}
\end{figure}

We further evaluated the proposed model using ExoNav II, a tendon-actuated concentric tube robot designed for follow-the-leader (FTL) navigation during spinal cord stimulation (SCS) procedures~\cite{moradkhani2025exonav}. The robot consists of a helically micromachined nitinol inner tube (see Fig.~\ref{fig:ExoNav_v2}(a)). This configuration enables ExoNav II to reach ventral and lateral spinal regions while preserving a minimally invasive dorsal entry pathway~\cite{moradkhani2025exonav}. For the purposes of this experiment, the outer tube remained passive, while the inner tube was actively actuated by a single tendon ($n = 1$, $m_1 = 1$).
To model this robot, the neutral axis is considered as the backbone, which is initially defined as a helix along the robot ($\bm{p_{c}}^*$). Furthermore, the initial state of the material orientation matrix ($R^*$) is defined as a rotation about the $\bm{Z}$-axis (which is pointed along the robot) and corresponding to the orientation of the notches. This way, the tendon route in the local frame is defined as a constant along arc length $s$, which leads to $\ddot{\bm{r}}=\dot{\bm{r}}=[0,0,0]^T$. Based on the defined $\bm{p_{c}}^*$ and $R^*$, the initial kinematic parameters $\bm{u}^*$ and $\bm{v}^*$ are obtained.
To clarify the distinction between this approach and the helical tendon routing scenario presented earlier, we consider, without loss of generality, the case where the robots are initially straight in their rest configuration. In our setup, the unactuated shape $\bm{p}^*$ is straight, and the associated material frame $R^*$ is the identity matrix. Under these conditions, the initial strain vectors $\bm{u}^*$ and $\bm{v}^*$ are given by $[0,\,0,\,0]^T$ and $\mathbb{I}_3$, respectively. Since the tendons follow a helical path, the routing vector $\bm{r}(s)$ varies along the arc length, resulting in nonzero $\dot{\bm{r}}$ and $\ddot{\bm{r}}$.
The model was evaluated using four discrete sensor placements along the inner tube. At each position, the tendon was actuated following the same predefined tension profile, and three frames during actuation were captured. The electromagnetic tracker (NDI Aurora) was used to record both the tip and intermediate positions of the inner tube during these trials. Because the same actuation profile was repeated at each sensor location, we used the average tendon tension over the captured frames as the actuation input to the model.
While the model captures the general trend of helical deformation (see Fig.~\ref{fig:ExoNav_v2}(b)), discrepancies were observed along the robot body. These deviations are attributed to the compliance introduced by the notched structure of the inner tube, which increases sensitivity to internal disturbances. Because the electromagnetic sensor was placed inside the robot during each measurement, it may have influenced the local shape—particularly near the sensor location. To mitigate this effect, only tip errors are reported, where the sensor's impact is minimal. Tip errors of $3.14$~mm, $4.15$~mm, and $3.59$~mm correspond to $4.16\%$, $5.50\%$, and $4.76\%$, respectively, relative to the $75.4$~mm robot length.

\section{Limitation and Future Work} \label{sec:Limit}
While the proposed model offers a general framework for simulating multi-tube TACTR robots, several limitations affect its practical applicability. The model is sensitive to initial conditions, and convergence may fail if the initial prediction is not sufficiently close to the true solution, particularly as the number of tubes or tendon tension increases. Convergence also depends on the smoothness of the tendon path; sharp or discontinuous routing can lead to numerical instability, requiring smooth, continuous paths for reliable results. Computation time increases significantly with system complexity, making the current implementation impractical for real-time control. Furthermore, the model does not account for hysteresis effects, leading to increased error during unloading phases observed in experiments.

Future work will focus on each of these limitations, improving the model’s robustness, computational efficiency, and suitability for real-time applications. Incorporating hysteresis effects such as friction, backlash, and material compliance will enhance accuracy during dynamic loading and unloading.

\section{Conclusion} \label{sec:conclusion}
This work presented a general modeling framework for estimating the quasi-static behavior of tendon-actuated concentric tube robots (TACTRs). Unlike existing models, the proposed approach accounts for shear and axial strain and supports arbitrary tendon routing paths. It is generalizable to $n$-tubes, and $m_i$ tendons (for tube-$i$) where $n$, and $m_i$ may take any non-negative integer values.
The model was validated through a series of experiments using a modular TACTR prototype in both two-tube and three-tube configurations. Trials included a variety of tendon routing conditions, such as static and helical paths. Additionally, experiments were conducted using previously developed robots, including a concentric tube robot without tendon actuation and other tendon-driven systems reported in the literature. In our experimental setup, the model achieved tip prediction errors within $4\%$ of the total robot length; larger errors were observed in other platforms.
We also examined how different tendon assignment strategies affect model accuracy. The model has several limitations, including sensitivity to initial conditions, high computation time, and the inability to capture hysteresis effects. Future efforts will focus on improving numerical stability, reducing runtime, and enabling real-time shape estimation and model-based control.

\section*{APPENDIX}
Here, we specify the remaining parameters in Eq.~(\ref{eq:finall_A}) and (\ref{eq:finall_b}).  The same procedure outlined in Section~\ref{subsec:ModelEquations} is applied to the \( \bm{d}_3 \) component of the moment equilibrium of the individual tubes, yielding the following equations for the first tube:
\begin{dmath} \label{eq:moment_localH}
{}_{1}H_u \cdot \dot{\bm{u}}_1 + {}_{1}H_v \cdot \dot{\bm{v}}_1 + {}_{1}H_{u_{d_3,i}} \mathbb{I}_3 \cdot \dot{u}_{d_3,i} + {}_{1}H_{\beta_{i}} \cdot \dot{\beta}_i = \bm{RHS}_{2,1}|_{\bm{d}_3}
\end{dmath}
\vspace{1em}
Where:
\small
\begin{dmath} \label{eq:Hu}
    {}_{1}H_u|_{\bm{d}_3} = K_{bt,1}+\sum_{j = 1}^{m_1}\frac{\lambda_{j,1}}{||\dot{\bm{p}}^b_{j,1}||^3}[\bm{r}_{j,1}][\dot{\bm{p}}^b_{j,1}]^2[\bm{r}_{j,1}]
\end{dmath}
\vspace{1em}
\begin{dmath} \label{eq:Hv}
    {}_{1}H_v|_{\bm{d}_3} = -\sum_{j = 1}^{m_1}\frac{\lambda_{j,1}}{||\dot{\bm{p}}^b_{j,1}||^3}[\bm{r}_{j,1}][\dot{\bm{p}}^b_{j,1}]^2
\end{dmath}
\vspace{1em}
\begin{dmath} \label{eq:Huz}
    {}_{1}H_{u_{d_3,i}}|_{\bm{d}_3} = 0
\end{dmath}
\vspace{1em}
\begin{dmath} \label{eq:Hbeta}
    {}_{1}H_{\beta_{i}}|_{\bm{d}_3} = 0
\end{dmath}
\vspace{1em}
\begin{dmath} \label{eq:RHS21}
    \bm{RHS}_{2,1}|_{\bm{d}_3} = K_{bt,1} \dot{\bm{u}}^*_i - [\bm{u}_1] K_{bt,1} (\bm{u}_1 - \bm{u}^*_1) - \\
    [\bm{v}_1] K_{se,1} (\bm{v}_1 - \bm{v}^*_1) + \sum_{j = 1}^{m_1} \frac{\lambda_{j,1}}{||\dot{\bm{p}}^b_{j,1}||^3} [\bm{r}_{j,1}] [\dot{\bm{p}}^b_{j,1}]^2 \left( [\bm{u}_1]\dot{\bm{p}}^b_{j,1} + [\bm{u}_1]\dot{\bm{r}}_{j,1} + \ddot{\bm{r}}_{j,1} \right)
\end{dmath}
\vspace{1em}
\normalsize
And for $i_{th}$ tube we have the following equations:
\begin{dmath} \label{eq:Hui}
    {}_{i}H_u|_{\bm{d}_3} = K_{bt,i}(R^T_{d_3}(\theta_i)-\mathbb{I}_{(3,3)})+\\
    \sum_{j = 1}^{m_i}\frac{\lambda_{j,i}}{||\dot{\bm{p}}^b_{j,i}||^3}[\bm{r}_{j,i}][\dot{\bm{p}}^b_{j,i}]^2[\bm{r}_{j,i}](R^T_{d_3}(\theta_i)-\mathbb{I}_{(3,3)})
\end{dmath}
\vspace{1em}
\begin{dmath} \label{eq:Hvi}
    {}_{i}H_v|_{\bm{d}_3} = -\sum_{j = 1}^{m_i} \frac{\lambda_{j,i}}{||\dot{\bm{p}}^b_{j,i}||^3} [\bm{r}_{j,i}][\dot{\bm{p}}^b_{j,i}]^2 \beta_i R^T_{d_3}(\theta_i)
\end{dmath}
\vspace{1em}
\begin{dmath} \label{eq:Huzi}
    {}_{i}H_{u_{d_3,i}}|_{\bm{d}_3} = K_{bt,i} + \sum_{j = 1}^{m_i} \frac{\lambda_{j,i}}{||\dot{\bm{p}}^b_{j,i}||^3} [\bm{r}_{j,i}] [\dot{\bm{p}}^b_{j,i}]^2 [\bm{r}_{j,i}]
\end{dmath}
\vspace{1em}
\begin{dmath} \label{eq:Hbetai}
    {}_{i}H_{\beta_{i}}|_{\bm{d}_3} = -\sum_{j = 1}^{m_i} \frac{\lambda_{j,i}}{||\dot{\bm{p}}^b_{j,i}||^3} [\bm{r}_{j,i}] [\dot{\bm{p}}^b_{j,i}]^2 R^T_{d_3}(\theta_i) \bm{v}_1
\end{dmath}
\vspace{1em}
\begin{dmath} \label{eq:RHS2i}
    \bm{RHS}_{2,i}|_{\bm{d}_3} = -K_{bt,i}(\dot{\theta}_i[\mathbb{I}_3]^T R^T_{d_3}(\theta_i)\bm{u}_1 - \dot{\bm{u}}^*_i)\\
    -[\bm{u}_i]K_{bt,i}(\bm{u}_i - \bm{u}^*_i) - [\bm{v}_i]K_{se,i}(\bm{v}_i - \bm{v}^*_i)\\
    + \sum_{j = 1}^{m_i} \frac{\lambda_{j,i}}{||\dot{\bm{p}}^b_{j,i}||^3}[\bm{r}_{j,i}][\dot{\bm{p}}^b_{j,i}]^2([\bm{u}_i]\dot{\bm{p}}^b_{j,i} + [\bm{u}_i]\dot{\bm{r}}_{j,i} + \ddot{\bm{r}}_{j,i})\\
    - \sum_{j = 1}^{m_i} \frac{\lambda_{j,i}}{||\dot{\bm{p}}^b_{j,i}||^3}[\bm{r}_{j,i}][\dot{\bm{p}}^b_{j,i}]^2[\bm{r}_{j,i}](\dot{\theta}_i[\mathbb{I}_3]^T R^T_{d_3}(\theta_i)\bm{u}_1)\\
    + \sum_{j = 1}^{m_i} \frac{\lambda_{j,i}}{||\dot{\bm{p}}^b_{j,i}||^3}[\bm{r}_{j,i}][\dot{\bm{p}}^b_{j,i}]^2 \beta_i [\mathbb{I}_3]^T R^T_{d_3}(\theta_i)\dot{\theta}_i \bm{v}_1
\end{dmath}
\vspace{1em}
\normalsize
Now turning to the sum of shear strain components of the equilibrium Eq.~(\ref{eq:equilibrium_eq_single_tube}), the same process can be repeated.
\begin{dmath} \label{eq:force_sum_local1}
    \sum_{i = 1}^n R_{d_3}(\theta_i)(\dot{\bm{n}}_i + [\bm{u}_i]\bm{n}_i + \bm{f}_t) = \mathbf{0} \big|_{\bm{d}_1, \bm{d}_2}
\end{dmath}
\vspace{1em}
\begin{dmath} \label{eq:force_sum_local4}
J_u \cdot \dot{\bm{u}}_1 + J_v \cdot \dot{\bm{v}}_1 + J_{u_{d_3,i}} \mathbb{I}_3 \cdot \dot{u}_{d_3,i} + J_{\beta_i} \cdot \dot{\beta}_i\\
= \bm{RHS}_3 \big|_{\bm{d}_1, \bm{d}_2}
\end{dmath}
\normalsize
\vspace{1em}
Where:
\begin{dmath} \label{eq:Ju}
    J_u|_{\bm{d}_1,\bm{d}_2} = \sum_{j = 1}^{m_1}\frac{\lambda_{j,1}}{||\dot{\bm{p}}^b_{j,1}||^3}[\dot{\bm{p}}^b_{j,1}]^2[\bm{r}_{j,1}]+\\
    \sum_{i = 2}^n\sum_{j = 1}^{m_i}\frac{\lambda_{j,i}}{||\dot{\bm{p}}^b_{j,i}||^3}R_{d_3}(\theta_i)[\dot{\bm{p}}^b_{j,i}]^2[\bm{r}_{j,i}](R^T_{d_3}(\theta_i)-\mathbb{I}_{(3,3)})
\end{dmath}
\vspace{1em}
\begin{dmath} \label{eq:Jv}
    J_v|_{\bm{d}_1,\bm{d}_2} =K_{se,1} - \sum_{j = 1}^{m_1} \frac{\lambda_{j,1}}{||\dot{\bm{p}}^b_{j,1}||^3} [\dot{\bm{p}}^b_{j,1}]^2 + \\
    \sum_{i = 2}^n R_{d_3}(\theta_i) K_{se,i} \beta_i R^T_{d_3}(\theta_i)-\\
    \sum_{j = 1}^{m_i} \frac{\lambda_{j,i}}{||\dot{\bm{p}}^b_{j,i}||^3} R_{d_3}(\theta_i) [\dot{\bm{p}}^b_{j,i}]^2 \beta_i R^T_{d_3}(\theta_i)
\end{dmath}
\vspace{1em}
\begin{dmath} \label{eq:Juz}
    J_{u_{d_3,i}}|_{\bm{d}_1,\bm{d}_2} = \sum_{j = 1}^{m_i} \frac{\lambda_{j,i}}{||\dot{\bm{p}}^b_{j,i}||^3} R_{d_3}(\theta_i) [\dot{\bm{p}}^b_{j,i}]^2 [\bm{r}_{j,i}]
\end{dmath}
\vspace{1em}
\begin{dmath} \label{eq:Jbeta}
    J_{\beta_i}|_{\bm{d}_1,\bm{d}_2} = R_{d_3}(\theta_i) K_{se,i} R^T_{d_3}(\theta_i) \bm{v}_1
    - \sum_{j = 1}^{m_i} \frac{\lambda_{j,i}}{||\dot{\bm{p}}^b_{j,i}||^3} R_{d_3}(\theta_i) [\dot{\bm{p}}^b_{j,i}]^2 R^T_{d_3}(\theta_i) \bm{v}_1
\end{dmath}
\vspace{1em}
\begin{dmath} \label{eq:RHS3}
    \bm{RHS}_3|_{\bm{d}_1,\bm{d}_2} = K_{se,1} \dot{\bm{v}}^*_1 - [\bm{u}_1] K_{se,1} (\bm{v}_1 - \bm{v}^*_1) + \sum_{j = 1}^{m_1} \frac{\lambda_{j,1}}{||\dot{\bm{p}}^b_{j,1}||^3} [\dot{\bm{p}}^b_{j,1}]^2 ([\bm{u}_1] \dot{\bm{p}}^b_{j,1} + [\bm{u}_1] \dot{\bm{r}}_{j,1} + \ddot{\bm{r}}_{j,1})\\
    - \sum_{i = 2}^n R_{d_3}(\theta_i) K_{se,i} (\beta_i [\mathbb{I}_3]^T R^T_{d_3}(\theta_i) \dot{\theta}_i \bm{v}_1 - \dot{\bm{v}}^*_i)\\
    - R_{d_3}(\theta_i) [\bm{u}_i] K_{se,i} (\bm{v}_i - \bm{v}^*_i)\\
    + \sum_{j = 1}^{m_i} \frac{\lambda_{j,i}}{||\dot{\bm{p}}^b_{j,i}||^3} R_{d_3}(\theta_i) [\dot{\bm{p}}^b_{j,i}]^2 ([\bm{u}_i] \dot{\bm{p}}^b_{j,i} + [\bm{u}_i] \dot{\bm{r}}_{j,i} + \ddot{\bm{r}}_{j,i})\\
    - \sum_{j = 1}^{m_i} \frac{\lambda_{j,i}}{||\dot{\bm{p}}^b_{j,i}||^3} R_{d_3}(\theta_i) [\dot{\bm{p}}^b_{j,i}]^2 [\bm{r}_{j,i}] (\dot{\theta}_i [\mathbb{I}_3]^T R^T_{d_3}(\theta_i) \bm{u}_1)\\
    + \sum_{j = 1}^{m_i} \frac{\lambda_{j,i}}{||\dot{\bm{p}}^b_{j,i}||^3} R_{d_3}(\theta_i) [\dot{\bm{p}}^b_{j,i}]^2 \beta_i [\mathbb{I}_3]^T R^T_{d_3}(\theta_i) \dot{\theta}_i \bm{v}_1
\end{dmath}
\vspace{1em}
\normalsize
Similar to the moment equations, the remaining scalar equations corresponding to the \( \bm{d}_3 \) component of the force equilibrium for each individual tube are given as follows:
\small
\begin{dmath} \label{eq:moment_localK}
{}_{1}K_u \cdot \dot{\bm{u}}_1 + {}_{1}K_v \cdot \dot{\bm{v}}_1 + {}_{1}K_{u_{d_3,i}} \mathbb{I}_3 \cdot \dot{u}_{d_3,i} + {}_{1}K_{\beta_i} \cdot \dot{\beta}_i = \bm{RHS}_{4,1}|_{\bm{d}_3}
\end{dmath}
\normalsize
\vspace{1em}
Where:
\small
\begin{dmath} \label{eq:Ku}
    {}_{1}K_u|_{\bm{d}_3} = \sum_{j = 1}^{m_1} \frac{\lambda_{j,1}}{||\dot{\bm{p}}^b_{j,1}||^3} [\dot{\bm{p}}^b_{j,1}]^2 [\bm{r}_{j,1}]
\end{dmath}
\vspace{1em}
\begin{dmath} \label{eq:Kv}
    {}_{1}K_v|_{\bm{d}_3} = K_{se,1} - \sum_{j = 1}^{m_1} \frac{\lambda_{j,1}}{||\dot{\bm{p}}^b_{j,1}||^3} [\dot{\bm{p}}^b_{j,1}]^2
\end{dmath}
\vspace{1em}
\begin{dmath} \label{eq:Kuz}
    {}_{1}K_{u_{d_3,i}}|_{\bm{d}_3} = 0
\end{dmath}
\vspace{1em}
\begin{dmath} \label{eq:Kbeta}
    {}_{1}K_{\beta_i}|_{\bm{d}_3} = 0
\end{dmath}
\vspace{1em}
\begin{dmath} \label{eq:RHS4_1}
    \bm{RHS}_{4,1}|_{\bm{d}_3} = K_{se,1} \dot{\bm{v}}^*_1 - [\bm{u}_1] K_{se,1} (\bm{v}_1 - \bm{v}^*_1) + \sum_{j = 1}^{m_1} \frac{\lambda_{j,1}}{||\dot{\bm{p}}^b_{j,1}||^3} [\dot{\bm{p}}^b_{j,1}]^2 ([\bm{u}_1] \dot{\bm{p}}^b_{j,1} + [\bm{u}_1] \dot{\bm{r}}_{j,1} + \ddot{\bm{r}}_{j,1})
\end{dmath}
\vspace{1em}
\normalsize
For $i_{th}$ tube we have the following equations:
\small
\begin{dmath} \label{eq:Kui}
    {}_{i}K_u|_{\bm{d}_3} = \sum_{j = 1}^{m_i} \frac{\lambda_{j,i}}{||\dot{\bm{p}}^b_{j,i}||^3} [\dot{\bm{p}}^b_{j,i}]^2 [\bm{r}_{j,i}] (R^T_{d_3}(\theta_i) - \mathbb{I}_{(3,3)})
\end{dmath}
\vspace{1em}
\begin{dmath} \label{eq:Kvi}
    {}_{i}K_v|_{\bm{d}_3} = K_{se,i} \beta_i R^T_{d_3}(\theta_i) - \sum_{j = 1}^{m_i} \frac{\lambda_{j,i}}{||\dot{\bm{p}}^b_{j,i}||^3} [\dot{\bm{p}}^b_{j,i}]^2 \beta_i R^T_{d_3}(\theta_i)
\end{dmath}
\vspace{1em}
\begin{dmath} \label{eq:Kuzi}
    {}_{i}K_{u_{d_3,i}}|_{\bm{d}_3} = \sum_{j = 1}^{m_i} \frac{\lambda_{j,i}}{||\dot{\bm{p}}^b_{j,i}||^3} [\dot{\bm{p}}^b_{j,i}]^2 [\bm{r}_{j,i}]
\end{dmath}
\vspace{1em}
\begin{dmath} \label{eq:Kbetai}
    {}_{1}K_{\beta_i}|_{\bm{d}_3} = K_{se,i} R^T_{d_3}(\theta_i) \bm{v}_1
    - \sum_{j = 1}^{m_i} \frac{\lambda_{j,i}}{||\dot{\bm{p}}^b_{j,i}||^3} [\dot{\bm{p}}^b_{j,i}]^2 R^T_{d_3}(\theta_i) \bm{v}_1
\end{dmath}
\vspace{1em}
\begin{dmath} \label{eq:RHS4i}
    \bm{RHS}_{4,i}|_{\bm{d}_3} = -K_{se,i} \left( \beta_i [\mathbb{I}_3]^T R^T_{d_3}(\theta_i) \dot{\theta}_i \bm{v}_1 - \dot{\bm{v}}^*_i \right)\\
    - [\bm{u}_i] K_{se,i} (\bm{v}_i - \bm{v}^*_i)\\
    + \sum_{j = 1}^{m_i} \frac{\lambda_{j,i}}{||\dot{\bm{p}}^b_{j,i}||^3} [\dot{\bm{p}}^b_{j,i}]^2 ([\bm{u}_i] \dot{\bm{p}}^b_{j,i} + [\bm{u}_i] \dot{\bm{r}}_{j,i} + \ddot{\bm{r}}_{j,i})\\
    - \sum_{j = 1}^{m_i} \frac{\lambda_{j,i}}{||\dot{\bm{p}}^b_{j,i}||^3} [\dot{\bm{p}}^b_{j,i}]^2 [\bm{r}_{j,i}] (\dot{\theta}_i [\mathbb{I}_3]^T R^T_{d_3}(\theta_i) \bm{u}_1)\\
    + \sum_{j = 1}^{m_i} \frac{\lambda_{j,i}}{||\dot{\bm{p}}^b_{j,i}||^3} [\dot{\bm{p}}^b_{j,i}]^2 \beta_i [\mathbb{I}_3]^T R^T_{d_3}(\theta_i) \dot{\theta}_i \bm{v}_1
\end{dmath}
\vspace{1em}
\normalsize
Finally, we can combine Eq.~(\ref{eq:moment_sum_local5}), (\ref{eq:moment_localH}), (\ref{eq:force_sum_local4}), and (\ref{eq:moment_localK}) to express our final differential equation as in Eq.~(\ref{eq:finall}).


\bibliographystyle{IEEEtran}
\bibliography{main}

\end{document}